\pdfoutput=1

\documentclass[11pt]{article}

\usepackage{eacl2023}

\usepackage{times}
\usepackage{latexsym}
\usepackage{graphicx}
\usepackage{float}
\usepackage[LGR, T1]{fontenc}

\usepackage[utf8]{inputenc}

\usepackage{microtype}

\usepackage{inconsolata}

\usepackage{array}
\usepackage{tabularx}
\usepackage{xcolor}

\usepackage[main=english,greek]{babel}
\usepackage{alphabeta}
\usepackage[utf8]{inputenc}

\newcommand{\el}{\selectlanguage{greek}}
\newcommand{\en}{\selectlanguage{english}}

\usepackage{multirow}
\usepackage{multicol}

\usepackage{geometry}

\title{GreekBART: The First Pretrained Greek Sequence-to-Sequence Model}

\author{Iakovos Evdaimon\textsuperscript{1} \And Hadi Abdine\textsuperscript{1} \And Christos Xypolopoulos\textsuperscript{1,2} \And Stamatis Outsios\textsuperscript{3} \AND Michalis Vazirgiannis\textsuperscript{1,4} \And Giorgos Stamou\textsuperscript{2}
\\\hspace{-5.5cm} \textsuperscript{1}\'Ecole Polytechnique, \textsuperscript{2}National Technical University of Athens, \\\hspace{-5.5cm}\textsuperscript{3}Athens University of Economics and Business, \textsuperscript{4}KTH Royal Institute of Technology}

\begin{document}
\selectlanguage{english}
\maketitle

\begin{abstract}
The era of transfer learning has revolutionized the fields of Computer Vision and Natural Language Processing, bringing powerful pretrained models with exceptional performance across a variety of tasks. Specifically, Natural Language Processing tasks have been dominated by transformer-based language models. In Natural Language Inference and Natural Language Generation tasks, the BERT model and its variants, as well as the GPT model and its successors, demonstrated exemplary performance. However, the majority of these models are pretrained and assessed primarily for the English language or on a multilingual corpus. In this paper, we introduce GreekBART, the first Seq2Seq model based on  BART-base architecture and pretrained on a large-scale Greek corpus. We evaluate and compare GreekBART against BART-random, Greek-BERT, and XLM-R on a variety of discriminative tasks. In addition, we examine its performance on two NLG tasks from GreekSUM, a newly introduced summarization dataset for the Greek language. The model, the code, and the new summarization dataset will be publicly available.
\end{abstract}


\section{Introduction and Related Work}
The field of machine learning has entered a new era with the establishment of transfer learning, providing new possibilities, especially in the areas of Computer Vision \cite{10.1145/3065386} and Natural Language Processing. Transfer learning has become a new trend that is so uncommon to train a model for computer vision or natural language processing tasks from scratch, dealing with the issue of insufficient training data for real-world machine learning applications. Tasks are solved by reusing pretrained models which are trained on enormous amounts of data, and the resulting models have reached state-of-the-art performance. Transformer \cite{vaswani2017attention} based pretrained models, as BERT \cite{devlin-etal-2019-bert} and its variants, are broadly used in Natural Language Processing, as have been shown to be effective in many tasks.

BART \cite{lewis-etal-2020-bart} is a denoising auto-encoder for pretraining sequence-to-sequence models. It is trained by corrupting text with an arbitrary noising function and learning a model to reconstruct the original text. It uses a standard Transformer-based neural machine translation architecture and a standard seq2seq architecture with a bidirectional encoder (like BERT) and a left-to-right decoder (like GPT \cite{Radford2018ImprovingLU}). This means the encoder's attention mask is fully visible, like BERT, and the decoder's attention mask is causal, like GPT2 \cite{Radford2019LanguageMA}. The unsupervised pretrained BART learns a language model, giving us the possibility to adapt it to a particular NLP task. So, large-scale labeled datasets are not required for fine-tuning. This type of model is suitable for machine translation, question-answering, and especially, text summarization tasks, but that does not mean that BART is insufficient in sequence classification tasks, on the contrary, it is also quite effective in that type of tasks.

In the last few years, a lot of research has been conducted on other languages, except for the English language. For instance,  CamemBERT \cite{Martin_2020} and BARThez \cite{kamal-eddine-etal-2021-barthez} for French language, CAMeLBERT \cite{inoue-etal-2021-interplay} and AraBART \cite{https://doi.org/10.48550/arxiv.2203.10945} for Arabic language, BART for Japanese language \cite{kim-komachi-2021-tmu}, BETO \cite{CaneteCFP2020} and NASes \cite{app11219872} for Spanish and Catalan languages, and BARTpho \cite{DBLP:journals/corr/abs-2109-09701} for Vietnamese language. Recently, a variety  of multilingual language models have been presented, covering multiple languages by being pretrained on a large-scale corpus of different languages, trying to learn the language model of multiple languages at once. Notably, M-BERT \cite{devlin-etal-2019-bert} is a case of a multilingual pretrained language model, which consists of the multilingual version of BERT, pretrained in the top 100 languages with the largest Wikipedias. Another case of a popular multilingual model is the XLM \cite{conneau2019cross} which is a transformer-based multilingual language model pretrained on Wikipedias of 15 languages. This model was trained in two auxiliary tasks, Masked Language Modeling, and the Translation Language Modeling task. Training a cross-lingual language model can be very beneficial for low-resource languages, as all languages are processed with the same shared vocabulary. \citealt{conneau-etal-2020-unsupervised} introduced XLM-R, an improved version of XLM based on the RoBERTa model. The model was trained with a cross-lingual masked language modeling objective on 2.5TB data in 100 languages from Common Crawl \cite{wenzek-etal-2020-ccnet,conneau-etal-2020-unsupervised}, increasing the amount of training available data for low-resource languages by two orders of magnitude on average. Finally, mBART \cite{liu-etal-2020-multilingual-denoising} is the multilingual version of BART and it is pretrained on a subset of 25 languages from the same dataset as XLM-R. In mBART, we use its 250K sentencepiece \cite{kudo-richardson-2018-sentencepiece} model which was trained using monolingual data for 100 languages from XLM-R, supporting languages beyond the original 25 mBART was trained on. The parameters of mBART25 are roughly 610M. Later, an extension of mBART in additional 25 languages (\emph{e.g.} total 50 languages) was proposed, mBART50 \cite{https://doi.org/10.48550/arxiv.2008.00401}, increasing the number of parameters to approximately 680M. Except for mBART and mBART50, all other aforementioned multilingual models support the Greek language. mBART25 and mBART50 are not pretrained on modern Greek, but it is included in their vocabulary. Nevertheless, multilingual models cannot compete with the performance of monolingual models in most NLP tasks. In the last months, another related model to BART that is in the spotlight of the NLP research area is ChatGPT \footnote{\url{https://openai.com/blog/chatgpt}}. ChatGPT is built on top of GPT-3 architecture\cite{brown2020language}, so it is a transformer-based language model that has been pretrained on massive amounts of text data and fine-tuned for conversational AI applications. Like BART, ChatGPT is capable of generating high-quality sequences of text, making it suitable for tasks such as text summarization and question answering. However, unlike BART, ChatGPT is specifically designed for conversational applications, making it well-suited for chatbots and other dialogue systems. In addition, ChatGPT's architecture is unidirectional, which means that it can generate text in a left-to-right sequence, making it more suitable for tasks such as language generation and dialogue.

Compared to languages that are widely spoken, Greek has fewer linguistic resources available. Especially, the available research in deep learning models for Greek is still very undeveloped. However, there are some efforts to develop datasets, models, knowledge bases, and frameworks for Greek NLP. \citealt{Outsios2018} presented the production of Greek word embeddings, where a large corpus of about 50GB (contains 120 million sentences), crawled from about 20 million URLs, was used for their work. Later, \citealt{lioudakis-etal-2020-ensemble} presented an ensemble method, Continuous Bag-of-Skip-grams, for extracting word representations for Greek. Recently, \citealt{Koutsikakis_2020} employed Greek-BERT, the first transformer-based language model, based on BERT, for the Greek language. The model was pretrained on a dataset of 29GB, achieving state-of-the-art performance in several NLP tasks in Greek. It is worth noting that \citealt{10.1145/3411408.3411410} have provided a throughout survey of the work that has been conducted in NLP for the Greek language.

In this contribution, we try to handle the issue that the multilingual models are not sufficient to compete with the monolingual ones and the limited available deep learning models for the Greek language. Thus, we propose the first pretrained Seq2Seq monolingual model for the Greek language. The model is called GreekBART, as we pretrained the BART-base architecture on a large monolingual Greek corpus. Despite the existence of the Greek-BERT \cite{Koutsikakis_2020}, our model exceeds the possibilities of Greek-BERT, focusing on generative tasks. GreekBART is evaluated on two different generative tasks and on four discriminative tasks. Our main contributions are:

\begin{itemize}
    \item We introduce the pretrained Seq2Seq model for the Greek language, based on BART-base architecture \cite{lewis-etal-2020-bart}, and pretrained on a large corpus of 87.6 GB. We examine the performance of our model in four discriminative tasks (\emph{i.e.} two classification tasks, one sentimental analysis task, and one Natural Language Inference task) and in two generative tasks.

    \item We present the first summarization dataset in Greek, GreekSUM, introducing two generative tasks and a classification task by processing this dataset.

    \item We compare GreekBART against popular language models, already pretrained or not on Greek. In the case of the discriminative tasks we collate our model, a BART-random model, Greek-BERT \cite{Koutsikakis_2020} and XLM-R \cite{conneau-etal-2020-unsupervised}. We also inspect the differences, in terms of performance, between  the GreekBART (\emph{i.e.} our model), BART-random model, mBART25 \cite{liu-etal-2020-multilingual-denoising} and mBART50 \cite{https://doi.org/10.48550/arxiv.2008.00401} on two novel generative tasks.

    \item We will publish our code and models\footnote{\url{https://github.com/iakovosevdaimon/GreekBART}}, providing access to everyone, who wants to further extend the applications of our work or take advantage of our contributions in favor of his/her work.
\end{itemize}


\section{GreekBART}
Our proposed model is based on BART \cite{lewis-etal-2020-bart} a denoising auto-encoder. We use the \emph{BASE} architecture, with 6 encoder and 6 decoder layers. Also, it is used 768 hidden dimensions, 12 attention heads in both the encoder and the decoder, and a normalization layer on top of both the encoder and the decoder \cite{liu-etal-2020-multilingual-denoising} is added. The purpose of these additional layers is to stabilize the training when FP16 precision \cite{DBLP:journals/corr/abs-1710-03740} is applied. The use of FP16 precision speeds up the pretraining of the model. In total, our model has roughly 181M parameters. Generally, we followed a similar methodology as \citealt{kamal-eddine-etal-2021-barthez}, in which a monolingual model in a different language than English is pretrained, following BART \cite{lewis-etal-2020-bart} and mBART \cite{liu-etal-2020-multilingual-denoising} methodologies.

\subsection{Pretraining corpus}
The pretrained corpus is produced by the following corpora: (a) the Greek part of Wikipedia\footnote{\url{https://dumps.wikimedia.org/elwiki/}}; (b) the Greek part of the European Parliament Proceedings Parallel Corpus (EuroParl)\footnote{\url{https://www.statmt.org/europarl/}} \cite{koehn-2005-europarl}; (c) the Greek part of OSCAR\footnote{\url{https://oscar-corpus.com/}} \cite{abadji-etal-2022-towards}, a clean version of CommonCrawl\footnote{\url{https://commoncrawl.org/}}; (d) the Greek Web Corpus, crawled from about 20 million Greek-language URLs\footnote{\url{http://nlp.polytechnique.fr/resources-greek}} \cite{Outsios2018}. In particular, we use the same datasets as the Greek-BERT \cite{Koutsikakis_2020} model, including also the dataset of \citealt{Outsios2018} in order to have a larger corpus that will be well suited for the pretraining of BART model. Moreover, by choosing these datasets we cover a wide variety of Greek language areas, which includes formal and informal text, news articles, encyclopedic information, and political conversations. This diverse range of text types helps to ensure that the pretraining of the BART model is robust and able to handle different styles and registers of Greek language use. Overall, the choice of datasets helps to ensure that the Greek BART model is well-equipped to handle a wide range of natural language processing tasks in the Greek language.

We preprocessed each of the aforementioned corpora by removing URLs, emojis, tags, and hashtags. Also, we erase comments, and some observed noisy sentences which do not provide any additional contextual meaning. The noisy sentences differ from dataset to dataset, so we had to detect them "manually". Furthermore, for all corpora except Wikipedia's dataset, we got rid of documents that contained less than one thousand characters. In the case of Wikipedia, we removed documents with less than thirty characters. Generally, we did not remove non-Greek characters, because we supposed that it will not prevent the GreekBART from understanding the language model, as their amount is insignificant. We deduplicated each corpora and then, we concatenated all of them in one corpus. Again, we deduplicated the merged dataset for a final time. The deduplication process was done using the runiq package\footnote{\url{https://github.com/whitfin/runiq}}. To generate our vocabulary, we used SentencePiece\footnote{\url{https://github.com/google/sentencepiece}} \cite{kudo-richardson-2018-sentencepiece} that implements byte-pair-encoding (BPE) \cite{sennrich-etal-2016-neural}. So, any type of pre-tokenization was not necessary. We fixed the size of the vocabulary to 50K sub-words and the SentencePiece model was trained on a 20GB random sample of the pretraining corpus. We set the character coverage to $99.95\%$. The total corpus size was 76.9/87.6GB before/after SentencePiece tokenization.

\begin{table}
\small
\centering
\begin{tabular}{|l|c|c|}
\hline
\multirow{2}{1cm}{\textbf{\small{Corpus}}} & \textbf{\small{Size before}} & \textbf{\small{Size after}} \\
    & \small{\textbf{deduplication}} & \small{\textbf{deduplication}} \\
    \hline \small{OSCAR} & $51.7$ & $44.6$\\
    \small{Greek Web Corpus} & $38.4$ & $30.9$\\
    \small{Wikipedia} & $0.9$ & $0.9$\\
    \small{EuroParl} & $0.5$ & $0.5$\\
    \hline
    \textbf{\small{Total}} & $91.5$ & $76.9$\\
    \hline
    \end{tabular}
    \caption{\label{tab:corpus}Datasets which consists of the GreekBART pretraining corpus (sizes in GB, before and after cleaning and deduplication).}
\end{table}

\subsection{Training details}
We adhere to the same pretraining process as BART. Thus, GreekBART tries to reconstruct the corrupted input by minimizing the cross-entropy loss between the decoder’s output and the original input. Two types of noise are applied in the input text. First, we employ the text infilling technique, where a number of text spans are replaced by a special token, called [MASK], masking $30 \%$ of text. A Poisson distribution with $(lambda=3.5)$ is used to determine the spans’ length. Sentence permutation is the second perturbation method, where the sentences of the input document are shuffled randomly.
 We pretrained GreekBART on Jean Zay, using a batch size equal to 768000 tokens per GPU, as we set the update frequency to 128. We used the Adam optimizer \cite{DBLP:journals/corr/KingmaB14} with $\epsilon = 10^{-6}, \beta_1 = 0.9$, and $\beta_2 = 0.999$, with a learning rate starting from $6.10^{-4}$ and decreasing linearly as a function of the training step. We used a warm-up of $6\%$ of the total number of training steps. In the first 12 epochs, we fixed the dropout to 0.1, for epochs 12 to 16 we decreased it to 0.05, and finally, we set it to zero for epochs 16 to 20. All experiments were carried out using the Fairseq library\footnote{\url{https://github.com/facebookresearch/fairseq}} \cite{ott-etal-2019-fairseq}.


\section{GreekSUM} \label{sec:greeksum}

\begin{table*}
\small
    \centering
    \begin{tabular}{|l|c|cc|cc|cc|} \hline 
    \multirow{2}{3cm}{Dataset} & \multirow{2}{3cm}{ train/val/test} & \multicolumn{2}{|c|}{ avg. doc length } & \multicolumn{2}{|c|}{ avg. summary length } & \multicolumn{2}{c|}{vocabulary size} \\
    & & words & sentences & words & sentences & docs & summaries \\
    \hline CNN & $90.3 / 1.22 / 1.09$ & $760.50$ & $33.98$ & $45.70$ & $3.58$ & $34$ & $89$ \\
    DailyMail & $197 / 12.15 / 10.40$ & $653.33$ & $29.33$ & $54.65$ & $3.86$ & $564$ & $180$ \\
    NY Times & $590 / 32.73 / 32.73$ & $800.04$ & $35.55$ & $45.54$ & $2.44$ & $1233$ & $293$ \\
    \hline XSum & $204 / 11.33 / 11.33$ & $431.07$ & $19.77$ & $23.26$ & $1.00$ & $399$ & $81$  \\
    OrangeSum Title & $30.6 / 1.5 / 1.5$ & $315.31$ & $10.87$ & $11.42$ & $1.00$ & $483$ & $43$ \\
    OrangeSum Abstract & $21.4 / 1.5 / 1.5$ & $350$ & $12.06$ & $32.12$ & $1.43$ & $420$ & $71$ \\
    GreekSUM Title & $146.046 / 10 / 10$ & $355.49$ & $14.26$ & $9.95$ & $1.05$ & $663$ & $91$ \\
    GreekSUM Abstract & $129.159 / 10 / 10$ & $368.97$ & $14.76$ & $24.55$ & $1.46$ & $629$ & $127$ \\
    \hline
    \end{tabular}
    \caption{Sizes (column 2) are given in thousands of documents. Document and summary lengths are in words, while vocabulary sizes are in thousands of tokens}
    \label{tab:greeksum_1}
\end{table*}

\begin{table*}
   \resizebox{1.\textwidth}{!}{
   \begin{tabular}{|l|cccc|ccc|ccc|}
        \hline \multirow{2}{4cm}{ Dataset } & \multicolumn{4}{|c|}{ \% of novel n-grams in gold summary } & \multicolumn{3}{c|}{ LEAD } & \multicolumn{3}{c|}{ EXT-ORACLE } \\
        & unigrams & bigrams & trigrams & 4-grams & R-1 & R-2 & R-L & R-1 & R-2 & R-L \\
        \hline CNN & $16.75$ & $54.33$ & $72.42$ & $80.37$ & $29.15$ & $11.13$ & $25.95$ & $50.38$ & $28.55$ & $46.58$ \\
        DailyMail & $17.03$ & $53.78$ & $72.14$ & $80.28$ & $40.68$ & $18.36$ & $37.25$ & $55.12$ & $30.55$ & $51.24$ \\
        NY Times & $22.64$ & $55.59$ & $71.93$ & $80.16$ & $31.85$ & $15.86$ & $23.75$ & $52.08$ & $31.59$ & $46.72$ \\
        \hline XSum & $35.76$ & $83.45$ & $95.50$ & $98.49$ & $16.30$ & $1.61$ & $11.95$ & $29.79$ & $8.81$ & $22.65$ \\
        OrangeSum Title & $26.54$ & $66.70$ & $84.18$ & $91.12$ & $19.84$ & $08.11$ & $16.13$ & $31.62$ & $17.06$ & $28.26$ \\
        OrangeSum Abstract & $30.03$ & $67.15$ & $81.94$ & $88.3$ & $22.21$ & $07.00$ & $15.48$ & $38.36$ & $20.87$ & $31.08$ \\
        
        GreekSUM Title & $26.7$ & $67.9$ & $84.5$ & $91.4$ & $14.68$ & $04.46$ & $14.37$ & $23.36$ & $07.39$ & $23.12$ \\
        GreekSUM Abstract & $20.6$ & $50.8$ & $65.3$ & $73.0$ & $17.11$ & $06.17$ & $16.69$ & $34.18$ & $14.17$ & $33.93$ \\
        \hline
    \end{tabular}}
    \caption{Degree of abstractivity of GreekSUM compared with that of other datasets. It depicts that GreekSUM follows XSum, and OrangeSum, being more abstractive than traditional summarization datasets.}
    \label{tab:greeksum_2}
\end{table*}

Transformer-based Seq2Seq models, including BART, can perform not only extractive but abstractive summarization, as well. This type of summarization is one of the most central and challenging evaluation tasks in NLP. However, there is not any available summarization dataset for the Greek language. Therefore, we created the first dataset in the Greek language, well-suited to the abstractive summarization task.

\subsection{Motivation}
Our main goal was to create a Greek version equivalent of the OrangeSum dataset\footnote{\url{https://github.com/Tixierae/OrangeSum}} \cite{kamal-eddine-etal-2021-barthez} and XSum dataset \cite{narayan-etal-2018-dont}. OrangeSum was produced by scraping articles, their single-sentence title, and their brief abstract from the "Orange Actu" website\footnote{\url{https://actu.orange.fr/}}. The title and the abstract of each article are written by the author of the article. Well-performed models on OrangeSum, as well as XSum, require a high degree of abstractivity.

\subsection{Data collection}
We followed a similar approach, scraping the "News24/7" website\footnote{\url{https://www.news247.gr/}}. News24/7 is one of the leading news websites in Greece, part of the 24 MEDIA digital publishing group\footnote{\url{https://www.24media.gr/}}. We collected data from web pages that span from October 2007 to June 2022, covering five major categories: politics, society, economy, culture, and world. Each article had a one-sentence title and a succinct abstract, features which were extracted, yielding two summarization tasks: GreekSUM Title and GreekSUM Abstract. The average length of these two novel tasks' gold summaries is 9.95 and 24.55 words respectively (see Table \ref{tab:greeksum_1}). 

\subsection{Post-processing}
Initially, we filtered the scrapped pages, removing all empty articles and articles whose titles were shorter than 2 words or whose abstracts were less than 5 words. Secondly, we filtered the duplicated articles (\emph{i.e.} articles with the same body, or with the same title, or with the same abstract), as an article can belong to more than one category, and thus be crawled multiple times. Finally, we noticed that several abstracts looked more like introductions rather than actual summaries of the article. Therefore, we eliminated 10\% of the articles with the highest proportion of novel unigrams in the abstracts. This corresponded to a threshold of 46.7\% novel unigrams. For both proposed summarization tasks, we reserved 10k pairs for testing, 10k for validation, and all the remaining pairs for training. The released GreekSUM dataset can be reproduced by using our code\footnote{\url{https://github.com/iakovosevdaimon/GreekSUM}}.

\subsection{Analysis}
In Table \ref{tab:greeksum_1} is compared the GreekSUM with OrangeSum, XSum, and the well-known CNN, DailyMail, and NY Times datasets \cite{NIPS2015_afdec700}. We can observe that GreekSUM and OrangeSum datasets are very equivalent in terms of average documents and summaries length. Also, GreekSUM has a similar scale to XSum. Inspecting the Table \ref{tab:greeksum_2}, it is noticeable that extractive methods (\emph{i.e.} LEAD and EXT-ORACLE) do not perform so well on GreekSUM, thus our dataset is less biased towards extractive models. Because of the poor performance of the two extractive methods, it seems that GreekSUM is more abstractive than the traditional summarization datasets (\emph{i.e.} CNN, DailyMail, NY Times). However, the summaries and the titles of GreekSUM do not display such a high degree of novelty as the ones of OrangeSum and XSum. In the GreekSUM dataset, there are 20.6\% novel unigrams in the abstracts and 26.7\% novel unigrams in the titles compared with 30\% in the OrangeSum Abstract, 26.5\% in the OrangeSum Title, and 35.7\% in XSum. Therefore, we can conclude that the summaries of GreekSUM are not as abstractive as we would like them to be.


\section{Experiments}
In this section, we present the results of all experiments. Basically, we have two types of downstream tasks, discriminative tasks, and summarization tasks. In the case of discriminative tasks, we compare GreekBART with BART-random, Greek-BERT \cite{Koutsikakis_2020}, and XLM-R model \cite{conneau-etal-2020-unsupervised}. Except for BART-random, the other models are already pretrained on the Greek language. So, we evaluate the performance of our model against the current state-of-the-art monolingual model pretrained only on the Greek language as well as against a widely used multilingual model. We fine-tuned all the above-mentioned models on the downstream tasks. 

For the summarization task, we set side by side the GreekBART, the BART-random and the two versions of mBART \cite{liu-etal-2020-multilingual-denoising,https://doi.org/10.48550/arxiv.2008.00401}. mBART25 and mBART50 are built upon the \emph{LARGE} architecture of BART, and they are pretrained on 25 and 50 languages respectively, excluding the Greek language. Therefore, we performed zero-shot learning for the summarization task. On the other hand, the BART-random model uses the same architecture and vocabulary as GreekBART, however, it is trained from scratch on the downstream tasks.

\subsection{Discriminative tasks}
Except for generative tasks, the BART model achieves remarkable results also in discriminative tasks \cite{lewis-etal-2020-bart}. In the case of sequence classification, a classification head is added on top of the model and the input is fed into both the encoder and the decoder. The representation of the final decoder token is used by the newly introduced multi-class linear classifier. We examine the performance of the models (\emph{i.e.} Greek-BERT, XLM-R, BART-random, GreekBART) on four discriminative tasks. More precisely, we evaluate our model on two classification tasks, one task of sentimental analysis and a Natural Language Inference (NLI) task.

\subsubsection{Training details}
In all experiments, we fine-tuned the models with a learning rate chosen from \{$10^{-4}, 5.10^{-5}, 10^{-5}$\}, based on the best validation score. We repeat each experiment 3 times with different seeds and we record the mean and standard deviation of their accuracy on the test set of each aforementioned task.

\subsubsection{NCC task (News Category Classification task)}
For the first classification task, we used the novel summarization dataset (GreekSum, see section \ref{sec:greeksum}) which we scraped from the news website News24/7 \footnote{\url{https://www.news247.gr/}}. We considered the five distinct subjects that an article may fall into politics, society, economy, culture, and world. These categories serve as labels for the classification task that our model is being trained to perform. Essentially, the model is fed with the content of an article and learns to predict which category it belongs to (\emph{i.e.} subject). We fine-tuned all examined models for 5 epochs, using a batch size equal to 32. For XLM-R model we set the learning rate equal to $5.10^{-5}$ while for the rest of the models, the learning rate is equal to $10^{-4}$. The training set consists of 146,046 samples, whereas both the validation and the test set have 10,000 instances exactly like the two summarization datasets (\emph{i.e.} GreekSUM Abstract and GreekSUM Title). 

In the second classification task, we used the proposed Greek classification dataset of \citealt{lioudakis-etal-2020-ensemble}, which was created from articles from Makedonia newspaper. The dataset contains 8005 articles from 18 different categories: Sports, Reportage, Economy, Politics, International, Television, Arts-Culture, Letters, Opinions, Interviews, Weather, Society, Advertisements, Biographies, Others, Articles, Police, and Zodiacs. We reserved 70\% of the dataset for train and the remaining 30\% for both validation and test. So, the train set consists of 5610 samples, whereas the test set and the validation set consist of 1191 and 1204 instances, respectively. All the models are fine-tuned for 20 epochs, with a batch size of 16 and a learning rate equal to $5.10^{-5}$. Due to the small size of the dataset, we trained the models for more epochs and smaller batch sizes.

\subsubsection{Natural Language Inference}
Cross-lingual Natural Language Inference Corpus (XNLI) \cite{conneau-etal-2018-xnli} contains pairs of sentences. The objective of this task is to determine whether the first sentence, also known as the premise, entails, contradicts, or is neutral in relation to the second sentence, referred to as the hypothesis. The XNLI corpus contains 5,000 test and 2,500 validation pairs, and 340k training pairs from the MultiNLI corpus \cite{williams-etal-2018-broad}. The dataset has been translated from English to 14 languages, including Greek. Unfortunately, a large number of the training pairs are of extremely poor quality, as they are produced by machine translation. This condition may affect the performance of models. We fine-tuned for 5 epochs, using 32 batches, and a learning rate equal to $5.10^{-5}$.

\subsubsection{Sentimental Analysis task}
We used a publicly available sentimental analysis dataset\footnote{\url{https://www.kaggle.com/datasets/nikosfragkis/greek-movies-dataset}} about movies' reviews in Greek. We preprocessed the dataset by mainly removing emojis and hashtags. Each instance consists of a review and a rating. To distinguish between positive and negative reviews, we established a threshold of 3 out of 5. Ratings above this threshold were categorized as positive reviews, while those at or below 3 out of 5 were classified as negative reviews. In an effort to create a balanced dataset, we aimed to include a similar number of positive and negative reviews. For the purpose of our task, we only retained the reviews and the ratings, discarding any additional information. We split the dataset into the train, validation, and test set. The train set consists of 104,157 samples, while the validation and test contain 22,320 and 22,318 instances respectively. We set the learning rate and the batch size equal to $5.10^{-5}$ and 16 respectively. We fine-tuned the models for 5 epochs. 

\begin{table*}
\small
   \resizebox{1.\textwidth}{!}{
    \begin{tabular}{|l|c|c|c|c|} 
        \hline \multirow{2}{0.5cm}{Model} &  \multicolumn{2}{|c|}{NCC} & \multirow{2}{1.5cm}{Sentimental Analysis} & \multirow{2}{1cm}{XNLI}\\
        \cline{2-3}
        & News24/7 (ours) & Makedonia \cite{lioudakis-etal-2020-ensemble} & & \\
        \hline
        Greek-BERT & $92.61^{\pm 0.19}$ & $89.45^{\pm 0.84}$ & $\textbf{86.39}^{\pm 0.06}$ & $78.6^{\pm 0.62}$ \\
        XLM-R  & $93.1^{\pm 0.51}$  & $89.6^{\pm 0.29}$ & $85.43^{\pm 0.05}$ & $78.2^{\pm 0.59}$\\
        BART-random & $91.33^{\pm 0.17}$ & $80.17^{\pm 0.09}$  & $80.87^{\pm 0.12}$  & $60.1^{\pm 0.43}$  \\
        GreekBART (ours) & $\textbf{93.2}^{\pm 0.29}$ & $\textbf{91.1}^{\pm 0.43}$  & $85.43^{\pm 0.19}$  & $\textbf{78.67}^{\pm 0.25}$  \\
        \hline
    \end{tabular}}
    \caption{Results on discriminative tasks. We present the mean accuracy as well as the standard deviation.}
    \label{tab:discr_1}
\end{table*}

\subsubsection{Results}
Table \ref{tab:discr_1} reports the test set accuracy on the four different tasks. We compare our model with Greek-BERT \cite{Koutsikakis_2020}, XLM-R \cite{conneau-etal-2020-unsupervised}, and BART-random. For all models, their corresponding \emph{BASE} architecture is used. Among the models, we observe that GreekBART is the best in almost all discriminative tasks, except for the sentimental analysis task, where Greek-BERT achieved the best performance. Generally, it is common for BERT models to perform better than BART models in that kind of tasks. The performance of our model (\emph{i.e.} GreekBART) verifies the results of BART paper \cite{lewis-etal-2020-bart} that models based on that architecture perform well on both generative and discriminative tasks.

\begin{table*}
\small
   \resizebox{1.\textwidth}{!}{
    \begin{tabular}{|cl|cccc|cccc|} 
        \hline \multirow{2}{0.5cm} & \multirow{2}{0.5cm} & \multicolumn{4}{|c|}{ GreekSUM Abstract } & \multicolumn{4}{|c|}{ GreekSUM Title }\\
        & & R-1 & R-2 & R-L & BertScore & R-1 & R-2 & R-L & BertScore \\
        \hline & LEAD & $17.11$ & $06.17$ & $16.69$ & $72.61/63.56$ & $14.68$ & $04.46$ & $14.37$ & $70/57.13$  \\
        & EXT-ORACLE & $34.18$ & $14.17$ & $33.93$ & $73.89/65.43$ & $23.36$ & $07.39$ & $23.12$ & $70.02/57.33$  \\
        \hline
        \multirow{2}{*}{\rotatebox[origin=c]{90}{\tiny{BASE}}} & BART-random & $13.85$ & $04.47$ & $13.65$ & $72.44/63.27$ & $11.55$ & $03.27$ & $11.42$ & $74.47/62.22$ \\
        & GreekBART (ours) & $\textbf{16.5}$ & $\textbf{06.13}$ & $\textbf{16.21}$ & $73.03/\textbf{64.46}$ & $15.35$ & $05.02$ & $15.18$ & $75.78/63.98$  \\
        \hline
        \multirow{2}{*}{\rotatebox[origin=c]{90}{\tiny{LARGE}}} & mBART25 & $15.07$ & $05.8$ & $14.82$ & $72.75/64.08$ & $16.09$ & $05.58$ & $15.93$ & $\textbf{76.81/65.38}$ \\
        & mBART50 & $15.53$ & $06.$ & $15.31$ & $\textbf{73.07}/64.43$ & $\textbf{16.1}$ & $\textbf{05.59}$ & $\textbf{15.96}$ & $\textbf{76.81/65.38}$ \\
        \hline
    \end{tabular}}
    \caption{Results on GreekSUM. Except for ROUGE, we provide also the BertScore. The left-hand BERTScore has calculated using the M-BERT model \cite{devlin-etal-2019-bert}, while the right-hand uses the Greek-BERT \cite{Koutsikakis_2020}}
    \label{tab:gen_1}
\end{table*}

\begin{table*}
\normalsize
   \resizebox{1.\textwidth}{!}{
    \begin{tabular}{|cl|ccccc|ccccc|} 
        \hline \multirow{2}{1cm} & \multirow{2}{1cm} & \multicolumn{5}{|c|}{ GreekSUM Abstract } & \multicolumn{5}{|c|}{ GreekSUM Title }\\
        & & unigrams & bigrams & trigrams & 4-grams & length & unigrams & bigrams & trigrams & 4-grams & length\\
        \hline & Gold & $20.6$ & $50.8$ & $65.3$ & $73.0$ & $24.55$ & $26.7$ & $67.9$ & $84.5$ & $91.4$ & $9.95$ \\
        \hline
        \multirow{2}{*}{\rotatebox[origin=c]{90}{\scriptsize{BASE}}} & BART-random & $9.6$ & $43.0$ & $64.5$ & $76.8$ & $20.27$ & $21.6$ & $69.4$ & $89.1$ & $95.8$ & $9.37$ \\
        & GreekBART (ours) & $\textbf{7.4}$ & $\textbf{23.5}$ & $\textbf{34.5}$ & $\textbf{42.2}$ & $23.63$ & $\textbf{14.9}$ & $\textbf{50.1}$ & $\textbf{69.3}$ & $\textbf{79.9}$ & $\textbf{9.78}$  \\
        \hline
        \multirow{2}{*}{\rotatebox[origin=c]{90}{\scriptsize{LARGE}}}& mBART25 & $6.2$ & $20.0$ & $29.4$ & $36.0$ & $26.22$ & $12.8$ & $46.6$ & $65.6$ & $76.2$ & $10.67$ \\
        & mBART50 & $6.5$ & $21.8$ & $32.3$ & $39.7$ & $23.95$ & $12.8$ & $46.6$ & $65.6$ & $76.2$ & $10.67$ \\
        \hline
    \end{tabular}}
    \caption{Proportion of novel n-grams in the generated summaries. Also, it is given the length (number of words) of the generated summaries}
    \label{tab:gen_2}
\end{table*}

\begin{table}
\small
\centering
\begin{tabular}{|c|l|c|}
\hline
& & \textbf{Repetitions (\%)}\\
\hline
\multirow{5}{*}{\rotatebox[origin=c]{90}{\normalsize{Abstract}}} & Gold & $7.77$ \\
& BART-random & $28.12$ \\ 
& GreekBART (ours) & $12.19$ \\
& mBART25 & $12.7$ \\ 
& mBART50 & $10.03$ \\
\hline
\end{tabular}
\begin{tabular}{|c|l|c|}
\hline
& & \textbf{Repetitions (\%)}\\
\hline
\multirow{5}{*}{\rotatebox[origin=c]{90}{\normalsize{Title}}} & Gold & $0.91$ \\
& BART-random & $8.76$ \\ 
& GreekBART (ours) & $3.62$ \\
& mBART25 & $2.52$ \\ 
& mBART50 & $2.52$ \\
\hline
\end{tabular}
\caption{The percentage of repeated words on the summaries}
\label{tab:repetitions}
\end{table}

\subsection{Summarization}
We evaluate our model in two distinct summarization tasks, in which the model learns to predict the title and the abstract of an article based on its corresponding content. In both generative tasks, the GreekBART was fine-tuned for 30 epochs with a learning rate equal to $5.10^{-5}$ that was warmed up for 6\% of the training steps and then decreased linearly to 0. We used the same set of hyper-parameters as those of GreekBART to train mBART25 and mBART50. While for BART-random, we trained the model for 60 epochs. To produce the summaries for the test set, we used ROUGE-L \cite{lin-2004-rouge} to select the checkpoint that was associated with the best validation score. In addition, we incorporated two extractive techniques as baselines: EXT-ORACLE and LEAD \cite{narayan-etal-2018-dont}. The LEAD technique generates a summary by extracting the first $N$ sentences from the document, with $N$ set to 1 in our case. On the other hand, EXT-ORACLE selects the set of sentences from the document that maximizes a specific score, with ROUGE-L being the score used in our implementation. In particular, we extracted the one sentence of the document with the highest ROUGE-L score. In Table \ref{tab:gen_1}, we report the ROUGE-1, ROUGE-2, ROUGE-L scores \cite{lin-2004-rouge} and two different BERTScores \cite{DBLP:journals/corr/abs-1904-09675}, using the M-BERT \cite{devlin-etal-2019-bert} model and the Greek-BERT model in order to calculate the contextual embeddings. BERTScore is a recently proposed metric that makes use of the contextual representations of the predicted and gold sentences. BERTScore focuses on semantic similarity between tokens of reference and hypothesis, trying to understand the meaning of what you have generated and what was supposed to be generated. We report BERTScore because ROUGE can mainly capture n-gram overlap, which is inadequate for the abstractive summarization setting. Some examples of the generated summarizations are available in the appendix section \ref{sec:appendix_abstract}, \ref{sec:appendix_title}.

\subsubsection{Quantitative results}

In Table \ref{tab:gen_1} we compare the performance of our models fine-tuned on the summarization task. Despite that GreekBART is a BART-\emph{BASE} model and it is compared with BART-\emph{LARGE} models, it is able to achieve better performance than all other models in the task of GreekSUM abstract. Only mBART50 achieves a slightly higher BERTScore than GreekBART when evaluated using the M-BERT model. On the other hand, both mBART models surpass our model in the GreekSUM title task. Although, even in that task the performance of GreekBART is comparable to one of the two mBART models, both in terms of ROUGE and BERTScore. Our evaluation indicates that mBART50 and GreekBART are the most promising models for the two summarization tasks. Specifically, mBART50 performs better overall in both generative tasks, being the top-performing model in the GreekSUM title task and second-best in the GreekSUM Abstract task, according to its ROUGE and BERTScores. On the other hand, GreekBART excels in the GreekSUM abstract task, but ranks third-best in the GreekSUM title task. Generally, it is remarkable the fact that both mBART models, which are not pretrained on the Greek language, are capable to achieve a good performance due to the size of GreekSUM dataset, which contains more than 100k training samples. It is clear that BART-random has the poorest performance by a significant margin. Finally, it is interesting that mBART50 has a better performance than mBART25 in terms of both ROUGE and BERTScore, while their only difference is the number of languages on which they are pretrained. This situation warrants further investigation, as it is possible that some of the additional 25 languages supported by mBART50 have roots in the Greek language, potentially contributing to a better understanding of the language model.

\subsubsection{Qualitative results}
As shown in Table \ref{tab:gen_2}, GreekBART is more abstractive than the two mBART models, as its generated summaries display a higher degree of novel n-grams. In general, none of the models surpass the LEAD method in terms of ROUGE scores. Furthermore, the ROUGE scores of the models suggest that the machine-generated summaries tend to be extractive, as the gold summaries are also predominantly extractive in nature. This situation is confirmed by the proportion of novel n-grams that are introduced (Table \ref{tab:gen_2}), where few new words are introduced in the gold summaries of GreekSUM, influencing, therefore, the training of the examined models, forcing them to generate more extractive summaries. Moreover, Table \ref{tab:gen_2} depicts that the length of all generated summaries is pretty close to the length of ground truth summaries. According to Table \ref{tab:repetitions} the generated summaries of mBART50 contain the smallest percentage of repetitions, with GreekBART following. The rate of repeated words on mBART50 summaries is close to the one of ground truth summaries. Finally, we notice that BART-random introduces many new words, however, they are irrelevant.

\begin{table}
\normalsize
\centering
\begin{tabular}{|c|l|c|}
\hline
& \textbf{System} & \textbf{Score}\\
\hline
& Gold & $45.24$ \\
\hline
\multirow{2}{*}{\rotatebox[origin=c]{90}{\scriptsize{BASE}}} & BART-random & $-72.62$ \\ 
& GreekBART (ours) & $10.71$ \\
\hline
\multirow{2}{*}{\rotatebox[origin=c]{90}{\scriptsize{LARGE}}} & mBART25 & $-03.57$ \\ 
& mBART50 & $\textbf{20.24}$ \\
\hline
\end{tabular}
\caption{The results of human evaluation study}
\label{tab:human_eval}
\end{table}

\subsubsection{Human Evaluation}
In order to further understand and validate the quantitative results, we conducted a human evaluation study, using Best-Worst Scaling \cite{louviere_flynn_marley_2015}. We chose 11 native Greek speakers from diverse age groups, ranging from 18 to 60 years old, with varying educational backgrounds and levels. Following \citealt{narayan-etal-2018-dont} method, we randomly selected 14 documents from the test set of GreekSUM abstract and for each document we generated all possible pairs of human-authored (Gold), GreekBART, BART-random, mBART25, and mBART50 summaries, resulting in a total of 140 pairs for all documents. Thus, each pair of summaries consists of two summaries generated by two different models. Volunteers were presented with a document and a pair of summaries and they should decide which one is the best summary and which was the worst, based on the accuracy (does the summary contain accurate facts?), the informativeness (is important information captured?) and the fluency (is the summary written in well-formed Greek?). Each summary pair was assigned randomly to three participants, and a system's score was determined by calculating the percentage of times it was selected as the \textit{best} summary, minus the percentage of times it was selected as the \textit{worst} summary. Thus, the maximum score that a model can achieve is $100$, whereas the minimum score can be $-100$. The results of the human evaluation study are presented in Table \ref{tab:human_eval}. Gold reaches first place, followed by mBART50 and GreekBART. According to the evaluators, Gold is by far the most preferred summary, while the score of mBART50 is remarkably higher than that of GreekBART, verifying our assumptions based on the quantitative results. Finally, the high negative score of BART-random indicates that its summaries were considered to be worse in the majority of cases.


\section{Conclusion}
We implemented GreekBART, the first pretrained Seq2Seq model for the Greek language specifically. Also, we created the first summarization dataset for the Greek language. Our model showed to outperform former state-of-the-art models on 3 out of 4 discriminative tasks and to be on par with BART-\emph{LARGE} models on summarization tasks. Moreover, we presented the capabilities of zero-shot learning, training from scratch a multilingual BART model on summarization tasks, even though it was not pretrained on the Greek language. As a future work, we can consider the creation of a more abstractive summarization dataset, and the investigation of any correlation between the Greek language and one or more of the 25 extra languages of mBART50. Finally, it would be interesting to try to boost the performance of mBART50 on summarization tasks by applying an affordable language-adaptive phase in order to further pretrain it on the Greek language for a logical number of epochs.


\section*{Ethics Statement}

The collection of the GreekSUM dataset was performed using a Python crawler that respected the \textit{robots.txt} of  \url{http://www.news247.gr}. As the dataset is used only for evaluation purposes the content follows the legal instructions listed on the webpage. 

For the training of GreekBART we used a cluster of GPUs consisting of 2 NVIDIA V100 GPUs for 20 days. As the majority of language models that are based on BART architecture the energy resources required for pretraining models currently are very high and need to be tackled soon \cite{strubell-etal-2019-energy}. 

\section*{Limitations}

The proposed GreekSUM dataset that we used for the evaluation of our model is limited to news articles from one webpage only. Thus, the capability of abstractive summarization of GreekBART is only assessed on one domain only. This is due to the fact that there is a lack of non-English benchmarks and tasks. This is also applicable in the discriminative tasks, where the only available ones for Greek are either sentence classification or natural language inference. While other evaluation datasets are not existing for the Greek language (i.e. Word Sense Disambiguation) or are not available to the public (i.e. Named Entity Recognition dataset).

On the other hand, GreekBART is only compared with extractive summarization methods or with large multi-lingual language models for the summarization task. Since it is the first base model for this language and since the base mBART model does not exist publicly, a fair in-depth comparison of GreekBART with other summarization systems could not be conducted. 

\section*{Acknowledgements}
This research was supported by the ANR chair AML/HELAS (ANR-CHIA-0020-01). 

This work was granted access to the HPC resources of IDRIS under the allocation 2022-AD011013750 made by GENCI.

We would like to express our sincere gratitude to all the participants who took part in this human evaluation study. Your time and effort in completing the questionnaires and participating in the study have been invaluable in helping us gather meaningful data.

Your willingness to share your experiences, insights, and opinions has been instrumental in informing our research, and we appreciate the trust you have placed in us. Your contributions have helped us improve our understanding of the topic under investigation and have the potential to make a significant impact on future research and practice.

We would also like to acknowledge the importance of obtaining informed consent from all participants before their involvement in the study. Your participation was entirely voluntary, and we appreciate your willingness to take part in the study.

Once again, we extend our sincere thanks to all the participants for their valuable contributions to this study.

\bibliography{anthology,custom}

\begin{thebibliography}{39}
\expandafter\ifx\csname natexlab\endcsname\relax\def\natexlab#1{#1}\fi

\bibitem[{Abadji et~al.(2022)Abadji, Ortiz~Suarez, Romary, and
  Sagot}]{abadji-etal-2022-towards}
Julien Abadji, Pedro Ortiz~Suarez, Laurent Romary, and Beno{\^\i}t Sagot. 2022.
\newblock \href {https://aclanthology.org/2022.lrec-1.463} {Towards a cleaner
  document-oriented multilingual crawled corpus}.
\newblock In \emph{Proceedings of the Thirteenth Language Resources and
  Evaluation Conference}, pages 4344--4355, Marseille, France. European
  Language Resources Association.

\bibitem[{Ahuir et~al.(2021)Ahuir, Hurtado, González, and
  Segarra}]{app11219872}
Vicent Ahuir, Lluís-F. Hurtado, José~Ángel González, and Encarna Segarra.
  2021.
\newblock \href {https://doi.org/10.3390/app11219872} {Nasca and nases: Two
  monolingual pre-trained models for abstractive summarization in catalan and
  spanish}.
\newblock \emph{Applied Sciences}, 11(21).

\bibitem[{Brown et~al.(2020)Brown, Mann, Ryder, Subbiah, Kaplan, Dhariwal,
  Neelakantan, Shyam, Sastry, Askell et~al.}]{brown2020language}
Tom Brown, Benjamin Mann, Nick Ryder, Melanie Subbiah, Jared~D Kaplan, Prafulla
  Dhariwal, Arvind Neelakantan, Pranav Shyam, Girish Sastry, Amanda Askell,
  et~al. 2020.
\newblock Language models are few-shot learners.
\newblock \emph{Advances in neural information processing systems},
  33:1877--1901.

\bibitem[{Cañete et~al.(2020)Cañete, Chaperon, Fuentes, Ho, Kang, and
  Pérez}]{CaneteCFP2020}
José Cañete, Gabriel Chaperon, Rodrigo Fuentes, Jou-Hui Ho, Hojin Kang, and
  Jorge Pérez. 2020.
\newblock Spanish pre-trained bert model and evaluation data.
\newblock In \emph{PML4DC at ICLR 2020}.

\bibitem[{Conneau et~al.(2020)Conneau, Khandelwal, Goyal, Chaudhary, Wenzek,
  Guzm{\'a}n, Grave, Ott, Zettlemoyer, and
  Stoyanov}]{conneau-etal-2020-unsupervised}
Alexis Conneau, Kartikay Khandelwal, Naman Goyal, Vishrav Chaudhary, Guillaume
  Wenzek, Francisco Guzm{\'a}n, Edouard Grave, Myle Ott, Luke Zettlemoyer, and
  Veselin Stoyanov. 2020.
\newblock \href {https://doi.org/10.18653/v1/2020.acl-main.747} {Unsupervised
  cross-lingual representation learning at scale}.
\newblock In \emph{Proceedings of the 58th Annual Meeting of the Association
  for Computational Linguistics}, pages 8440--8451, Online. Association for
  Computational Linguistics.

\bibitem[{Conneau and Lample(2019)}]{conneau2019cross}
Alexis Conneau and Guillaume Lample. 2019.
\newblock Cross-lingual language model pretraining.
\newblock \emph{Advances in neural information processing systems}, 32.

\bibitem[{Conneau et~al.(2018)Conneau, Rinott, Lample, Williams, Bowman,
  Schwenk, and Stoyanov}]{conneau-etal-2018-xnli}
Alexis Conneau, Ruty Rinott, Guillaume Lample, Adina Williams, Samuel Bowman,
  Holger Schwenk, and Veselin Stoyanov. 2018.
\newblock \href {https://doi.org/10.18653/v1/D18-1269} {{XNLI}: Evaluating
  cross-lingual sentence representations}.
\newblock In \emph{Proceedings of the 2018 Conference on Empirical Methods in
  Natural Language Processing}, pages 2475--2485, Brussels, Belgium.
  Association for Computational Linguistics.

\bibitem[{Devlin et~al.(2019)Devlin, Chang, Lee, and
  Toutanova}]{devlin-etal-2019-bert}
Jacob Devlin, Ming-Wei Chang, Kenton Lee, and Kristina Toutanova. 2019.
\newblock \href {https://doi.org/10.18653/v1/N19-1423} {{BERT}: Pre-training of
  deep bidirectional transformers for language understanding}.
\newblock In \emph{Proceedings of the 2019 Conference of the North {A}merican
  Chapter of the Association for Computational Linguistics: Human Language
  Technologies, Volume 1 (Long and Short Papers)}, pages 4171--4186,
  Minneapolis, Minnesota. Association for Computational Linguistics.

\bibitem[{Eddine et~al.(2022)Eddine, Tomeh, Habash, Roux, and
  Vazirgiannis}]{https://doi.org/10.48550/arxiv.2203.10945}
Moussa~Kamal Eddine, Nadi Tomeh, Nizar Habash, Joseph~Le Roux, and Michalis
  Vazirgiannis. 2022.
\newblock \href {https://doi.org/10.48550/ARXIV.2203.10945} {Arabart: a
  pretrained arabic sequence-to-sequence model for abstractive summarization}.

\bibitem[{Hermann et~al.(2015)Hermann, Kocisky, Grefenstette, Espeholt, Kay,
  Suleyman, and Blunsom}]{NIPS2015_afdec700}
Karl~Moritz Hermann, Tomas Kocisky, Edward Grefenstette, Lasse Espeholt, Will
  Kay, Mustafa Suleyman, and Phil Blunsom. 2015.
\newblock \href
  {https://proceedings.neurips.cc/paper/2015/file/afdec7005cc9f14302cd0474fd0f3c96-Paper.pdf}
  {Teaching machines to read and comprehend}.
\newblock In \emph{Advances in Neural Information Processing Systems},
  volume~28. Curran Associates, Inc.

\bibitem[{Inoue et~al.(2021)Inoue, Alhafni, Baimukan, Bouamor, and
  Habash}]{inoue-etal-2021-interplay}
Go~Inoue, Bashar Alhafni, Nurpeiis Baimukan, Houda Bouamor, and Nizar Habash.
  2021.
\newblock \href {https://aclanthology.org/2021.wanlp-1.10} {The interplay of
  variant, size, and task type in {A}rabic pre-trained language models}.
\newblock In \emph{Proceedings of the Sixth Arabic Natural Language Processing
  Workshop}, pages 92--104, Kyiv, Ukraine (Virtual). Association for
  Computational Linguistics.

\bibitem[{Kamal~Eddine et~al.(2021)Kamal~Eddine, Tixier, and
  Vazirgiannis}]{kamal-eddine-etal-2021-barthez}
Moussa Kamal~Eddine, Antoine Tixier, and Michalis Vazirgiannis. 2021.
\newblock \href {https://doi.org/10.18653/v1/2021.emnlp-main.740} {{BART}hez: a
  skilled pretrained {F}rench sequence-to-sequence model}.
\newblock In \emph{Proceedings of the 2021 Conference on Empirical Methods in
  Natural Language Processing}, pages 9369--9390, Online and Punta Cana,
  Dominican Republic. Association for Computational Linguistics.

\bibitem[{Kim and Komachi(2021)}]{kim-komachi-2021-tmu}
Hwichan Kim and Mamoru Komachi. 2021.
\newblock \href {https://doi.org/10.18653/v1/2021.wat-1.13} {{TMU} {NMT} system
  with {J}apanese {BART} for the patent task of {WAT} 2021}.
\newblock In \emph{Proceedings of the 8th Workshop on Asian Translation
  (WAT2021)}, pages 133--137, Online. Association for Computational
  Linguistics.

\bibitem[{Kingma and Ba(2015)}]{DBLP:journals/corr/KingmaB14}
Diederik~P. Kingma and Jimmy Ba. 2015.
\newblock \href {http://arxiv.org/abs/1412.6980} {Adam: {A} method for
  stochastic optimization}.
\newblock In \emph{3rd International Conference on Learning Representations,
  {ICLR} 2015, San Diego, CA, USA, May 7-9, 2015, Conference Track
  Proceedings}.

\bibitem[{Koehn(2005)}]{koehn-2005-europarl}
Philipp Koehn. 2005.
\newblock \href {https://aclanthology.org/2005.mtsummit-papers.11} {{E}uroparl:
  A parallel corpus for statistical machine translation}.
\newblock In \emph{Proceedings of Machine Translation Summit X: Papers}, pages
  79--86, Phuket, Thailand.

\bibitem[{Koutsikakis et~al.(2020)Koutsikakis, Chalkidis, Malakasiotis, and
  Androutsopoulos}]{Koutsikakis_2020}
John Koutsikakis, Ilias Chalkidis, Prodromos Malakasiotis, and Ion
  Androutsopoulos. 2020.
\newblock \href {https://doi.org/10.1145/3411408.3411440} {Greek-bert: The
  greeks visiting sesame street}.
\newblock In \emph{11th Hellenic Conference on Artificial Intelligence}, SETN
  2020, page 110–117, New York, NY, USA. Association for Computing Machinery.

\bibitem[{Krizhevsky et~al.(2017)Krizhevsky, Sutskever, and
  Hinton}]{10.1145/3065386}
Alex Krizhevsky, Ilya Sutskever, and Geoffrey~E. Hinton. 2017.
\newblock \href {https://doi.org/10.1145/3065386} {Imagenet classification with
  deep convolutional neural networks}.
\newblock \emph{Commun. ACM}, 60(6):84–90.

\bibitem[{Kudo and Richardson(2018)}]{kudo-richardson-2018-sentencepiece}
Taku Kudo and John Richardson. 2018.
\newblock \href {https://doi.org/10.18653/v1/D18-2012} {{S}entence{P}iece: A
  simple and language independent subword tokenizer and detokenizer for neural
  text processing}.
\newblock In \emph{Proceedings of the 2018 Conference on Empirical Methods in
  Natural Language Processing: System Demonstrations}, pages 66--71, Brussels,
  Belgium. Association for Computational Linguistics.

\bibitem[{Lewis et~al.(2020)Lewis, Liu, Goyal, Ghazvininejad, Mohamed, Levy,
  Stoyanov, and Zettlemoyer}]{lewis-etal-2020-bart}
Mike Lewis, Yinhan Liu, Naman Goyal, Marjan Ghazvininejad, Abdelrahman Mohamed,
  Omer Levy, Veselin Stoyanov, and Luke Zettlemoyer. 2020.
\newblock \href {https://doi.org/10.18653/v1/2020.acl-main.703} {{BART}:
  Denoising sequence-to-sequence pre-training for natural language generation,
  translation, and comprehension}.
\newblock In \emph{Proceedings of the 58th Annual Meeting of the Association
  for Computational Linguistics}, pages 7871--7880, Online. Association for
  Computational Linguistics.

\bibitem[{Lin(2004)}]{lin-2004-rouge}
Chin-Yew Lin. 2004.
\newblock \href {https://aclanthology.org/W04-1013} {{ROUGE}: A package for
  automatic evaluation of summaries}.
\newblock In \emph{Text Summarization Branches Out}, pages 74--81, Barcelona,
  Spain. Association for Computational Linguistics.

\bibitem[{Lioudakis et~al.(2020)Lioudakis, Outsios, and
  Vazirgiannis}]{lioudakis-etal-2020-ensemble}
Michalis Lioudakis, Stamatis Outsios, and Michalis Vazirgiannis. 2020.
\newblock \href {https://aclanthology.org/2020.loresmt-1.13} {{A}n ensemble
  method for producing word representations focusing on the {G}reek language}.
\newblock In \emph{Proceedings of the 3rd Workshop on Technologies for MT of
  Low Resource Languages}, pages 99--107, Suzhou, China. Association for
  Computational Linguistics.

\bibitem[{Liu et~al.(2020)Liu, Gu, Goyal, Li, Edunov, Ghazvininejad, Lewis, and
  Zettlemoyer}]{liu-etal-2020-multilingual-denoising}
Yinhan Liu, Jiatao Gu, Naman Goyal, Xian Li, Sergey Edunov, Marjan
  Ghazvininejad, Mike Lewis, and Luke Zettlemoyer. 2020.
\newblock \href {https://doi.org/10.1162/tacl_a_00343} {Multilingual denoising
  pre-training for neural machine translation}.
\newblock \emph{Transactions of the Association for Computational Linguistics},
  8:726--742.

\bibitem[{Louviere et~al.(2015)Louviere, Flynn, and
  Marley}]{louviere_flynn_marley_2015}
Jordan~J. Louviere, Terry~N. Flynn, and A.~A.~J. Marley. 2015.
\newblock \href {https://doi.org/10.1017/CBO9781107337855} {\emph{Best-Worst
  Scaling: Theory, Methods and Applications}}.
\newblock Cambridge University Press.

\bibitem[{Martin et~al.(2020)Martin, Muller, Su{\'{a} }rez, Dupont, Romary,
  de~la Clergerie, Seddah, and Sagot}]{Martin_2020}
Louis Martin, Benjamin Muller, Pedro Javier~Ortiz Su{\'{a} }rez, Yoann Dupont,
  Laurent Romary, {\'{E}}ric de~la Clergerie, Djam{\'{e}} Seddah, and
  Beno{\^{\i}}t Sagot. 2020.
\newblock \href {https://doi.org/10.18653/v1/2020.acl-main.645} {{CamemBERT}: a
  tasty french language model}.
\newblock In \emph{Proceedings of the 58th Annual Meeting of the Association
  for Computational Linguistics}. Association for Computational Linguistics.

\bibitem[{Micikevicius et~al.(2017)Micikevicius, Narang, Alben, Diamos, Elsen,
  Garc{\'{\i}}a, Ginsburg, Houston, Kuchaiev, Venkatesh, and
  Wu}]{DBLP:journals/corr/abs-1710-03740}
Paulius Micikevicius, Sharan Narang, Jonah Alben, Gregory~F. Diamos, Erich
  Elsen, David Garc{\'{\i}}a, Boris Ginsburg, Michael Houston, Oleksii
  Kuchaiev, Ganesh Venkatesh, and Hao Wu. 2017.
\newblock \href {http://arxiv.org/abs/1710.03740} {Mixed precision training}.
\newblock \emph{CoRR}, abs/1710.03740.

\bibitem[{Narayan et~al.(2018)Narayan, Cohen, and
  Lapata}]{narayan-etal-2018-dont}
Shashi Narayan, Shay~B. Cohen, and Mirella Lapata. 2018.
\newblock \href {https://doi.org/10.18653/v1/D18-1206} {Don{'}t give me the
  details, just the summary! topic-aware convolutional neural networks for
  extreme summarization}.
\newblock In \emph{Proceedings of the 2018 Conference on Empirical Methods in
  Natural Language Processing}, pages 1797--1807, Brussels, Belgium.
  Association for Computational Linguistics.

\bibitem[{Ott et~al.(2019)Ott, Edunov, Baevski, Fan, Gross, Ng, Grangier, and
  Auli}]{ott-etal-2019-fairseq}
Myle Ott, Sergey Edunov, Alexei Baevski, Angela Fan, Sam Gross, Nathan Ng,
  David Grangier, and Michael Auli. 2019.
\newblock \href {https://doi.org/10.18653/v1/N19-4009} {fairseq: A fast,
  extensible toolkit for sequence modeling}.
\newblock In \emph{Proceedings of the 2019 Conference of the North {A}merican
  Chapter of the Association for Computational Linguistics (Demonstrations)},
  pages 48--53, Minneapolis, Minnesota. Association for Computational
  Linguistics.

\bibitem[{Outsios et~al.(2018)Outsios, Skianis, Meladianos, Xypolopoulos, and
  Vazirgiannis}]{Outsios2018}
Stamatis Outsios, Konstantinos Skianis, Polykarpos Meladianos, Christos
  Xypolopoulos, and Michalis Vazirgiannis. 2018.
\newblock Word embeddings from large-scale greek web content.
\newblock \emph{arXiv preprint arXiv:1810.06694}.

\bibitem[{Papantoniou and Tzitzikas(2020)}]{10.1145/3411408.3411410}
Katerina Papantoniou and Yannis Tzitzikas. 2020.
\newblock \href {https://doi.org/10.1145/3411408.3411410} {Nlp for the greek
  language: A brief survey}.
\newblock In \emph{11th Hellenic Conference on Artificial Intelligence}, SETN
  2020, page 101–109, New York, NY, USA. Association for Computing Machinery.

\bibitem[{Radford et~al.(2018)Radford, Narasimhan, Salimans, Sutskever
  et~al.}]{Radford2018ImprovingLU}
Alec Radford, Karthik Narasimhan, Tim Salimans, Ilya Sutskever, et~al. 2018.
\newblock Improving language understanding by generative pre-training.

\bibitem[{Radford et~al.(2019)Radford, Wu, Child, Luan, Amodei, Sutskever
  et~al.}]{Radford2019LanguageMA}
Alec Radford, Jeffrey Wu, Rewon Child, David Luan, Dario Amodei, Ilya
  Sutskever, et~al. 2019.
\newblock Language models are unsupervised multitask learners.
\newblock \emph{OpenAI blog}, 1(8):9.

\bibitem[{Sennrich et~al.(2016)Sennrich, Haddow, and
  Birch}]{sennrich-etal-2016-neural}
Rico Sennrich, Barry Haddow, and Alexandra Birch. 2016.
\newblock \href {https://doi.org/10.18653/v1/P16-1162} {Neural machine
  translation of rare words with subword units}.
\newblock In \emph{Proceedings of the 54th Annual Meeting of the Association
  for Computational Linguistics (Volume 1: Long Papers)}, pages 1715--1725,
  Berlin, Germany. Association for Computational Linguistics.

\bibitem[{Strubell et~al.(2019)Strubell, Ganesh, and
  McCallum}]{strubell-etal-2019-energy}
Emma Strubell, Ananya Ganesh, and Andrew McCallum. 2019.
\newblock \href {https://doi.org/10.18653/v1/P19-1355} {Energy and policy
  considerations for deep learning in {NLP}}.
\newblock In \emph{Proceedings of the 57th Annual Meeting of the Association
  for Computational Linguistics}, pages 3645--3650, Florence, Italy.
  Association for Computational Linguistics.

\bibitem[{Tang et~al.(2020)Tang, Tran, Li, Chen, Goyal, Chaudhary, Gu, and
  Fan}]{https://doi.org/10.48550/arxiv.2008.00401}
Yuqing Tang, Chau Tran, Xian Li, Peng-Jen Chen, Naman Goyal, Vishrav Chaudhary,
  Jiatao Gu, and Angela Fan. 2020.
\newblock \href {https://doi.org/10.48550/ARXIV.2008.00401} {Multilingual
  translation with extensible multilingual pretraining and finetuning}.

\bibitem[{Tran et~al.(2021)Tran, Le, and
  Nguyen}]{DBLP:journals/corr/abs-2109-09701}
Nguyen~Luong Tran, Duong~Minh Le, and Dat~Quoc Nguyen. 2021.
\newblock \href {http://arxiv.org/abs/2109.09701} {Bartpho: Pre-trained
  sequence-to-sequence models for vietnamese}.
\newblock \emph{CoRR}, abs/2109.09701.

\bibitem[{Vaswani et~al.(2017)Vaswani, Shazeer, Parmar, Uszkoreit, Jones,
  Gomez, Kaiser, and Polosukhin}]{vaswani2017attention}
Ashish Vaswani, Noam Shazeer, Niki Parmar, Jakob Uszkoreit, Llion Jones,
  Aidan~N Gomez, {\L}ukasz Kaiser, and Illia Polosukhin. 2017.
\newblock Attention is all you need.
\newblock \emph{Advances in neural information processing systems}, 30.

\bibitem[{Wenzek et~al.(2020)Wenzek, Lachaux, Conneau, Chaudhary, Guzm{\'a}n,
  Joulin, and Grave}]{wenzek-etal-2020-ccnet}
Guillaume Wenzek, Marie-Anne Lachaux, Alexis Conneau, Vishrav Chaudhary,
  Francisco Guzm{\'a}n, Armand Joulin, and Edouard Grave. 2020.
\newblock \href {https://aclanthology.org/2020.lrec-1.494} {{CCN}et: Extracting
  high quality monolingual datasets from web crawl data}.
\newblock In \emph{Proceedings of the 12th Language Resources and Evaluation
  Conference}, pages 4003--4012, Marseille, France. European Language Resources
  Association.

\bibitem[{Williams et~al.(2018)Williams, Nangia, and
  Bowman}]{williams-etal-2018-broad}
Adina Williams, Nikita Nangia, and Samuel Bowman. 2018.
\newblock \href {https://doi.org/10.18653/v1/N18-1101} {A broad-coverage
  challenge corpus for sentence understanding through inference}.
\newblock In \emph{Proceedings of the 2018 Conference of the North {A}merican
  Chapter of the Association for Computational Linguistics: Human Language
  Technologies, Volume 1 (Long Papers)}, pages 1112--1122, New Orleans,
  Louisiana. Association for Computational Linguistics.

\bibitem[{Zhang et~al.(2019)Zhang, Kishore, Wu, Weinberger, and
  Artzi}]{DBLP:journals/corr/abs-1904-09675}
Tianyi Zhang, Varsha Kishore, Felix Wu, Kilian~Q. Weinberger, and Yoav Artzi.
  2019.
\newblock \href {http://arxiv.org/abs/1904.09675} {Bertscore: Evaluating text
  generation with {BERT}}.
\newblock \emph{CoRR}, abs/1904.09675.

\end{thebibliography}
\bibliographystyle{acl_natbib}

\clearpage
\newgeometry{margin=2.5cm}

\appendix
\section*{Appendices}
\section{Appendix-GreekSUM Abstract}
\label{sec:appendix_abstract}

In this appendix section, we present the reference and model summaries of 5 randomly selected documents from the test set of the GreekSUM Abstract.

    \begin{table}[h!] 
     \centering 
     \begin{tabular}{|cr|p{5.5in}|}
     \hline 
     & Document & \el «Ο κύβος ερρίφθη. Ο 'Αρμιν Λάσετ θα είναι ο υποψήφιος Καγκελάριος για την Χριστιανική Ένωση», δήλωσε πριν από λίγο ο Αρχηγός της Χριστιανοκοινωνικής Ένωσης \selectlanguage{english}(CSU) \selectlanguage{greek}και Πρωθυπουργός της Βαυαρίας Μάρκους Ζέντερ, αναγνωρίζοντας το αποτέλεσμα της ψηφοφορίας του προεδρείου του Χριστιανοδημοκρατικού Κόμματος\selectlanguage{english} (CDU), \selectlanguage{greek}το οποίο σε ποσοστό 77,5\% τάχθηκε υπέρ της υποψηφιότητας του κ. Λάσετ. Πριν από την συνεδρίαση του προεδρείου του \selectlanguage{english}CDU, \selectlanguage{greek}ο κ. Ζέντερ είχε δηλώσει ότι παραχωρεί στο \selectlanguage{english} CDU \selectlanguage{greek}το προβάδισμα στην επιλογή του υποψήφιου Καγκελάριου της Χριστιανικής Ένωσης\selectlanguage{english} (CDU/CSU) \selectlanguage{greek}και σήμερα επανέλαβε ότι δέχεται το αποτέλεσμα «χωρίς μηνισικακία» και ότι τάσσεται υπέρ της ενότητας της Χριστιανικής Ένωσης. \\ 
     \hline 
     \hline 
     \multirow{5}{*}[-3.5em]{\rotatebox[origin=c]{90}{\textsc{Abstract}}} 
     & Gold &  \selectlanguage{greek}Ο Άρμιν Λάσετ θα είναι ο υποψήφιος των \selectlanguage{english}CDU \selectlanguage{greek}και \selectlanguage{english}CSU \elγια την καγκελαρία της Γερμανίας στις εκλογές του Σεπτεμβρίου.\\ 
     & BART-random &\selectlanguage{greek}Ο Αρμιν Λάσετ θα είναι ο υποψήφιος πρωθυπουργός της Χριστιανικής Ένωσης, μετά από σχετική συνεδρίαση.\\ 
     & mBART25 &\selectlanguage{greek}Ο πρωθυπουργός της Βαυαρίας δέχθηκε το αποτέλεσμα της ψηφοφορίας του προεδρείου του \selectlanguage{english}CDU, \selectlanguage{greek}το οποίο σε ποσοστό 77,5\% τάχθηκε υπέρ της υποψηφιότητας του Αρμιν Λάσετ. \\
     & mBART50 &\selectlanguage{greek}Σε ποσοστό 77,5\% τάχθηκε υπέρ της υποψηφιότητας του Αρμιν Λάσετ στο προεδρείο του \selectlanguage{english}CDU, \elο Πρωθυπουργός της Βαυαρίας Μάρκους Ζέντερ.\\
     & GreekBART &\selectlanguage{greek}Υπέρ του Αρμιν Λάσετ τάσσεται ο Μάρκους Ζέντερ, αναγνωρίζοντας το αποτέλεσμα της ψηφοφορίας του προεδρείου του \selectlanguage{english}CDU.\\ 
     \hline 
     \end{tabular} 
     \caption{Example 1-GreekSUM Abstract} 
     \label{appendix:table1abs} 
     \end{table}

\clearpage

 \begin{table}[h!] 
     \centering 
     \begin{tabular}{|cr|p{5.5in}|}
     \hline 
     & Document & \el Κλειστή είναι η λεωφόρος Βασ. Κωνσταντίνου στο ύψος του Παναθηναϊκού Σταδίου, με αποτέλεσμα να έχει δημιουργηθεί κυκλοφοριακό πρόβλημα, καθώς έχει χυθεί μεγάλη ποσότητα λαδιού από φορτηγό, στην συμβολή με την λεωφόρο Βασ. Όλγας και είναι επικίνδυνη η διέλευση των οχημάτων. Η Τροχαία έχει διακόψει την κυκλοφορία στο κατερχόμενο ρεύμα στο ύψος της οδού Ριζάρη και στο ανερχόμενο από την αρχή της Αρδητού και κάνει εκτροπή, αλλά έχει δημιουργηθεί μποτιλιάρισμα. Στο σήμειο που έχουν χυθεί τα λάδια βρίσκονται συνεργεία του Δήμου, που ρίχνουν πριονίδι και άλλα υλικά για να αντιμετωπίσουν την ολοσθηρότητα του οδοστρώματος και να αποκατασταθεί η κυκλοφορία. \\ 
     \hline 
     \hline 
     \multirow{5}{*}[-3.5em]{\rotatebox[origin=c]{90}{\textsc{Abstract}}} & Gold &  \selectlanguage{greek}Η Τροχαία έχει διακόψει την κυκλοφορία στο κατερχόμενο ρεύμα στο ύψος της οδού Ριζάρη και στο ανερχόμενο από την αρχή της Αρδητού και κάνει εκτροπή - Μποτιλιάρισμα στο σημείο.\\ 
     & BART-random &\selectlanguage{greek}Η Τροχαία, που έχει δημιουργηθεί στο ύψος του Παναθηναϊκού, έχει διακοπεί την κυκλοφορία των οχημάτων στην λεωφόρο Βασ.Α.\\ 
     & mBART25 &\selectlanguage{greek}Κλειστή είναι η λεωφόρος Βασ. Κωνσταντίνου στο ύψος του Παναθηναϊκού Σταδίου, με αποτέλεσμα να έχει δημιουργηθεί κυκλοφοριακό πρόβλημα. \\
     & mBART50 &\selectlanguage{greek}Κυκλοφοριακό πρόβλημα στο ύψος του Παναθηναϊκού Σταδίου, καθώς έχει χυθεί μεγάλη ποσότητα λαδιού από φορτηγό σε λεωφόρο Βασ. Κωνσταντίνου.\\
     & GreekBART &\selectlanguage{greek}Κυκλοφοριακό πρόβλημα έχει δημιουργηθεί στην Λεωφόρο Βασ. Κωνσταντίνου στο ύψος του Παναθηναϊκού Σταδίου, με αποτέλεσμα να έχει δημιουργηθεί μποτιλιάρισμα.\\ 
     \hline 
     \end{tabular} 
     \caption{Example 2-GreekSUM Abstract} 
     \label{appendix:table2abs} 
     \end{table}

\clearpage
 
  \begin{table}[h!] 
     \centering 
     \begin{tabular}{|cr|p{5.5in}|}
     \hline 
     & Document & \el Η Καγκελάριος Άνγκελα Μέρκελ δεν θα παραστεί στην επίσημη δεξίωση που θα παραθέσει την Παρασκευή ο Ομοσπονδιακός Πρόεδρος Φρανκ-Βάλτερ Σταϊνμάιερ προς τιμήν του Προέδρου της Τουρκίας Ρετζέπ Ταγίπ Ερντογάν, σύμφωνα με κυβερνητικές πηγές τις οποίες επικαλείται το περιοδικό \en «Der Spiegel». \el Η δεξίωση αλλά και οι στρατιωτικές τιμές με τις οποίες θα υποδεχθεί τον προσκεκλημένο του ο Γερμανός Πρόεδρος προκαλούν σοβαρές αντιδράσεις στον πολιτικό κόσμο της χώρας. Η Μέρκελ είναι πάντα προσκεκλημένη του Ομοσπονδιακού Προέδρου σε δεξιώσεις ή επίσημα δείπνα που παρατίθενται προς τιμήν υψηλών προσκεκλημένων. Η ίδια ωστόσο συνηθίζει να παρευρίσκεται μόνο σε εξαιρετικές περιπτώσεις. Η τελευταία φορά που παρέστη σε κάτι ανάλογο ήταν το επίσημο δείπνο που είχε παρατεθεί το 2015 προς τιμήν της Βασίλισσας Ελισάβετ, ενώ την προηγούμενη χρονιά είχε παρευρεθεί στο δείπνο με τον Εμίρη του Κατάρ. Αντιθέτως, δεν είχε παρευρεθεί στην δεξίωση προς τιμή του Κινέζου Προέδρου Σι Τζινπίνγκ το 2017. Η Καγκελάριος όμως δεν θα είναι η μόνη που θα απορρίψει την πρόσκληση του Σταϊνμάιερ. Ο Πρόεδρος των Φιλελευθέρων \en(FDP) \elΚρίστιαν Λίντντερ ανακοίνωσε ότι δεν σκοπεύει να παραστεί, καθώς δεν επιθυμεί «να συμμετάσχει στην προπαγάνδα του Ερντογάν». Την ίδια στάση θα τηρήσει και η εκπρόσωπος του κόμματος για την εξωτερική πολιτική, Μπιτζάν Ντζιρ-Σαράι, ενώ σύσσωμη η ηγετική ομάδα των Πρασίνων, οι συμπρόεδροι Αναλένα Μπέρμποκ και Ρόμπερτ Χάμπεκ και οι επικεφαλής της Κοινοβουλευτικής Ομάδας Κάτριν Γκέρινγκ-Έκαρτ και 'Αντον Χοφράιτερ, δήλωσαν ότι θα απέχουν από την δεξίωση. Το ίδιο ισχύει και για τους επικεφαλής της Εναλλακτικής για την Γερμανία \en(AfD) \el'Αλεξάντερ Γκάουλαντ και Αλίς Βαϊντέλ και για την επικεφαλής της Κ. Ο. της Αριστεράς Σεβίμ Νταγκντελέν. Αντιθέτως, την πρόθεσή του να παραστεί στην δεξίωση στο Προεδρικό Ανάκτορο \en Bellevue \el εξέφρασε ο πρώην Πρόεδρος των Πρασίνων Τζεμ Έζντεμιρ, διευκρινίζοντας ταυτόχρονα ότι ο Τούρκος Πρόεδρος «δεν είναι κανονικός Πρόεδρος και δεν αξίζει» να παρατεθεί δεξίωση προς τιμήν του. Με την παρουσία του, δήλωσε ο κ. Έζντεμιρ στην \en «Tagesspiegel», \el ελπίζει να στείλει ένα μήνυμα τόσο προς την Τουρκία όσο και προς την τουρκογερμανική κοινότητα: «Η αντιπολίτευση στην Γερμανία είναι μέρος της πολιτικής αυτής της χώρας, είμαστε ένα σταθερό και απαραίτητο συστατικό στοιχείο της δημοκρατίας μας. Ο κ. Ερντογάν θα πρέπει να με ανεχθεί». \\ 
     \hline 
     \hline 
     \multirow{5}{*}[-3.5em]{\rotatebox[origin=c]{90}{\textsc{Abstract}}} & Gold &  \selectlanguage{greek}Η καγκελάριος είναι πάντα προσκεκλημένη του ομοσπονδιακού προέδρου σε δεξιώσεις ή δείπνα προς τιμήν υψηλών προσκεκλημένων, ωστόσο δίνει το παρών μόνο σε εξαιρετικές περιπτώσεις.\\ 
     & BART-random &\selectlanguage{greek}Δεν θα παραστεί στην επίσημη δεξίωση που θα παραθέσει την Τουρκία προς τιμήν του Ρετζέπ Ταγίπ Ερντογάν ο εκπρόσωπος της Γερμανίας Άνγκελα Μέρκελ.\\ 
     & mBART25 &\selectlanguage{greek}Αντιδράσεις από τον πολιτικό κόσμο της χώρας προκαλούν η δεξίωση που θα παραθέσει ο Φρανκ-Βάλτερ Σταϊνμάιερ προς τιμήν του Προέδρου της Τουρκίας - Δεν θα είναι η μόνη που θα απορρίψει την πρόσκληση του Σταϊνμάιερ. \\
     & mBART50 &\selectlanguage{greek}Η Μέρκελ είναι πάντα προσκεκλημένη του Ομοσπονδιακού Προέδρου σε δεξιώσεις ή επίσημα δείπνα που παρατίθενται προς τιμήν υψηλών προσκεκλημένων. Η ίδια ωστόσο συνηθίζει να παρευρίσκεται μόνο σε εξαιρετικές περιπτώσεις.\\
     & GreekBART &\selectlanguage{greek}Από τον πολιτικό κόσμο της Γερμανίας. Η Άνγκελα Μέρκελ δεν θα παραστεί στην επίσημη δεξίωση προς τιμήν του Γερμανού Προέδρου Φρανκ-Βάλτερ Σταϊνμάιερ.\\ 
     \hline 
     \end{tabular} 
     \caption{Example 3-GreekSUM Abstract} 
     \label{appendix:table3abs} 
     \end{table}

\clearpage

      \begin{table}[h!] 
     \centering 
     \begin{tabular}{|cr|p{5.5in}|}
     \hline 
     & Document & \el Από το 2011 και μετά αρκετοί εκατοντάδες άνθρωποι έχουν πεθάνει στην προσπάθειά τους να βγάλουν την τέλεια \en selfie. \el Οι περισσότεροι θάνατοι έχουν λάβει χώρα στην Ινδία. Ακολουθεί η Ρωσία, οι Ηνωμένες Πολιτείες και ύστερα το Πακιστάν με τους νεκρούς συνολικά να φτάνουν τους 259. Βέβαια υπάρχουν κάποια σημεία, τα οποία σύμφωνα με έρευνες, παρουσιάζουν μεγαλύτερη επικινδυνότητα, όπως το νερό και οι ψηλές κυλιόμενες σκάλες. Οι πιο «συνηθισμένες» αιτίες θανάτου από \en selfie \el συμπεριλαμβάνουν τον πνιγμό, την πτώση, τη σύγκρουση με κινούμενο όχημα και τις φωτιές. Όσον αφορά τα στατιστικά στοιχεία τα 3/4 των θυμάτων είναι άνδρες και κάτω από την ηλικία των 30. Αν και οι γυναίκες βγάζουν περισσότερες \en selfie \el σύμφωνα με τις μελέτες, οι άνδρες είναι πιο επιρρεπείς στον κίνδυνο. Ακόμα, οι τουρίστες είναι αυτοί που πλήττονται πιο συχνά στην προσπάθεια να βγάλουν μια φωτογραφία που θα εντυπωσιάσει τους ακολούθους τους. Οι αρχές ψάχνουν τρόπους προκειμένου να αποτρέψουν τους θανάτους. Για παράδειγμα η ρωσική αστυνομία μοίρασε φυλλάδια, τα οποία εμπεριείχαν προειδοποιήσεις σχετικά με τους κινδύνους που «καραδοκούν» πίσω από μια \en selfie. \el Στις Ηνωμένες Πολιτείες, τα εθνικά πάρκα έχουν εκδώσει οδηγούς για το πώς να βγάζεις «ασφαλείς» \en selfies, \el ενώ στην Ινδία υπάρχουν επίσημα σχεδιασμένες πινακίδες που προειδοποιούν για υψηλού κινδύνου περιοχές ή αλλιώς \en “No selfie zones”.\el Αν και η εμμονή με τις \en selfie \el δεν φαίνεται να περνάει οι αρχές κάνουν ότι μπορούν για να περιορίσουν την επικινδυνότητα και τους θανάτους. \\ 
     \hline 
     \hline 
     \multirow{5}{*}[-3.5em]{\rotatebox[origin=c]{90}{\textsc{Abstract}}} & Gold &  \selectlanguage{greek}Οι πιο «συνηθισμένες» αιτίες θανάτου από selfie συμπεριλαμβάνουν πνιγμό, πτώση, και τη σύγκρουση με κινούμενο όχημα - Άνδρες κάτω των 30 τα περισσότερα θύματα. \\ 
     & BART-random &\selectlanguage{greek}Οι Ηνωμένες Πολιτείες, Ινδία, Αν. και Πακιστάν και Αν. Ινδία αναζητούν αναζητούν στοιχεία για να βγάλουν την τέλεια \en selfie \el τους στην προσπάθειά τους.\\ 
     & mBART25 &\selectlanguage{greek}Η Ινδία μετράει τους 259 θανάτους από \en selfie,\el τα οποία συμπεριλαμβάνουν τον πνιγμό, την πτώση, τη σύγκρουση με κινούμενο όχημα και τις φωτιές. Οι αρχές ψάχνουν τρόπους προκειμένου να αποτρέψουν τους θανάτους. \\
     & mBART50 &\selectlanguage{greek}Στην Ινδία, τα εθνικά πάρκα έχουν εκδώσει οδηγούς για το πώς να βγάζεις «ασφαλείς» \en selfies, \el ενώ στην Ινδία υπάρχουν επίσημα σχεδιασμένες πινακίδες που προειδοποιούν για υψηλού κινδύνου περιοχές.\\
     & GreekBART &\selectlanguage{greek}Πολλοί άνθρωποι έχουν πεθάνει στην προσπάθειά τους να βγάλουν μια \en selfie, \el με τις «συνηθισμένες» αιτίες να συμπεριλαμβάνουν τον πνιγμό, την πτώση, τη σύγκρουση με κινούμενο όχημα και τις φωτιές.\\ 
     \hline 
     \end{tabular} 
     \caption{Example 4-GreekSUM Abstract} 
     \label{appendix:table4abs} 
     \end{table}

\clearpage

     \begin{table}[h!] 
     \centering 
     \begin{tabular}{|cr|p{5.5in}|}
     \hline 
     & Document & \el Στην απώλεια του Μίκη Θεοδωράκη αναφέρθηκε ο πρωθυπουργός Κυριάκος Μητσοτάκης στην έναρξη της συνεδρίασης του Υπουργικού Συμβουλίου, κηρύσσοντας τριήμερο εθνικό πένθος. Ο πρωθυπουργός ειδικότερα δήλωσε: “Τη σημερινή μας συνεδρίαση σκιάζει δυστυχώς μία πολύ θλιβερή είδηση: Ο Μίκης Θεοδωράκης περνά πια στην αιωνιότητα. Η φωνή του σίγησε και μαζί του σίγησε και ολόκληρος ο Ελληνισμός. Όπως είχε γραφτεί και για τον Παλαμά, «όλοι είχαμε ξεχάσει πως είναι θνητός». Όμως, μας αφήνει παρακαταθήκη τα τραγούδια του, την πολιτική του δράση, αλλά και την εθνική του προσφορά σε κρίσιμες στιγμές. Η Ρωμιοσύνη σήμερα κλαίει. Και γι’ αυτό και με απόφαση της κυβέρνησης από σήμερα κηρύσσεται τριήμερο εθνικό πένθος. Όπως ξέρετε, είχα την τιμή να τον γνωρίζω για πολλά χρόνια και σχετικά πρόσφατα μάλιστα τον είχα επισκεφτεί. Οι συμβουλές του ήταν πάντα πολύτιμες για μένα, κυρίως αυτές που αφορούσαν στην ενότητα του λαού μας και στην υπέρβαση των διαχωριστικών γραμμών. Πιστεύω πως η καλύτερη τιμή προς αυτόν τον παγκόσμιο Έλληνα θα είναι εμείς, με το καθημερινό μας έργο, να κάνουμε πράξη αυτό ακριβώς το μήνυμά του. Ο Μίκης είναι η Ιστορία μας και πρέπει να τη συνεχίσουμε όπως θα ήθελε και εκείνος.” Πέθανε ο Μίκης Θεοδωράκης - Ορφάνεψε η Ρωμιοσύνη Έλενα Ακρίτα - Ο ‘Ηλιος (που κρύφτηκε) και ο Χρόνος (που χάθηκε), Μίκη Μίκης Θεοδωράκης: Τα 5 τραγούδια του σπουδαίου μουσικού που «μιλούν» στην ψυχή της Ελλάδας Ο πολιτικός Μίκης Θεοδωράκης: Πάντα στο πλευρό των απλών ανθρώπων. \\ 
     \hline 
     \hline 
     \multirow{5}{*}[-3.5em]{\rotatebox[origin=c]{90}{\textsc{Abstract}}} & Gold &  \selectlanguage{greek}Η Ρωμιοσύνη σήμερα κλαίει δήλωσε ο πρωθυπουργός στην έναρξη της συνεδρίασης του υπουργικού συμβουλίου αναφερόμενος στο θάνατο του Μίκη Θεοδωράκη. \\ 
     & BART-random &\selectlanguage{greek}Ο πρωθυπουργός κατά την έναρξη της συνεδρίασης του Υπουργικού Συμβουλίου κηρύσσοντας την απώλεια του Μίκη Θεοδωράκη.\\ 
     & mBART25 &\selectlanguage{greek}Ο πρωθυπουργός Κυριάκος Μητσοτάκης απο το υπουργικό συμβούλιο για τον θάνατο του Μίκη Θεοδωράκη. \\
     & mBART50 &\selectlanguage{greek}Τριήμερο εθνικό πένθος κηρύχθηκε στη συνεδρίαση του υπουργικού συμβουλίου, με τον πρωθυπουργό να σημειώνει ότι ο Μίκης Θεοδωράκης περνά πια στην αιωνιότητα.\\
     & GreekBART &\selectlanguage{greek}Το δικό του μήνυμα για την απώλεια του Μίκη Θεοδωράκη έστειλε ο πρωθυπουργός Κυριάκος Μητσοτάκης κατά τη συνεδρίαση του Υπουργικού Συμβουλίου.\\ 
     \hline 
     \end{tabular} 
     \caption{Example 5-GreekSUM Abstract} 
     \label{appendix:table5abs} 
     \end{table}

\clearpage

     \section{Appendix- GreekSUM Title}
    \label{sec:appendix_title}
\begin{flushleft}
In the second section of the appendices, we present the reference and model titles of 5 randomly selected documents from the test set of the GreekSUM Title.
\end{flushleft}

    \begin{table}[h!] 
     \centering 
     \begin{tabular}{|cr|p{5.5in}|}
     \hline 
     & Document & \el Ένας 33χρονος έχασε τη ζωή του, ύστερα από σύγκρουση δύο αυτοκίνητων, έξω από τη Θεσσαλονίκη. Όπως έγινε γνωστό, το θανατηφόρο τροχαίο συνέβη στις 2.15 μετά τα μεσάνυχτα σε παράδρομο της Εγνατίας Οδού, στο ύψος του Ωραιοκάστρου. Σύμφωνα με την Αστυνομία, ο 33χρονος, οδηγός του ενός οχήματος, διακομίστηκε στο νοσοκομείο Παπαγεωργίου, όπου όμως λίγη αργότερα υπέκυψε στα τραύματά του, ενώ η οδηγός του άλλου οχήματος υπέστη ελαφρά τραύματα. Οι ακριβείς συνθήκες υπό τις οποίες προκλήθηκε η σύγκρουση ερευνώνται από το αρμόδιο τμήμα τροχαίας. \\ 
     \hline 
     \hline 
     \multirow{5}{*}[-3.5em]{\rotatebox[origin=c]{90}{\textsc{Title}}} & Gold &  \selectlanguage{greek}Τροχαίο δυστύχημα στη Θεσσαλονίκη με έναν νεκρό\\ 
     & BART-random &\selectlanguage{greek}Τροχαίο έξω από τη Θεσσαλονίκη - Δύο τραυματίες\\ 
     & mBART25 &\selectlanguage{greek}Θεσσαλονίκη: Νεκρός 33χρονος ύστερα από σύγκρουση δύο αυτοκίνητων \\
     & mBART50 &\selectlanguage{greek}Θεσσαλονίκη: Νεκρός 33χρονος ύστερα από σύγκρουση δύο αυτοκίνητων\\
     & GreekBART &\selectlanguage{greek}Τροχαίο στη Θεσσαλονίκη: Νεκρός 33χρονος σε παράδρομο\\ 
     \hline 
     \end{tabular} 
     \caption{Example 1-GreekSUM Title} 
     \label{appendix:table1t} 
     \end{table}

\clearpage

 \begin{table}[h!] 
     \centering 
     \begin{tabular}{|cr|p{5.5in}|}
     \hline 
     & Document & \el Ολες οι χώρες της Ευρωπαϊκής Ενωσης συμφωνούν ότι δεν θα πληρώσουν τη Ρωσία απευθείας σε ρούβλια για τις εισαγωγές ρωσικού φυσικού αερίου, δήλωσαν υψηλόβαθμοι ευρωπαίοι αξιωματούχοι, σημειώνοντας ότι οι επόμενες πληρωμές είναι προγραμματισμένες για τις 20 Μαΐου. «Αυτό που γνωρίζουμε, και υπάρχει συναίνεση επ΄αυτού μεταξύ όλων των κρατών μελών, είναι ότι κανείς δεν είναι πρόθυμος να πληρώσει σε ρούβλια», δήλωσε ο ένας αξιωματούχος κατά την διάρκεια ενημέρωσης των δημοσιογράφων και προσθέτοντας ότι η Ευρωπαϊκή Επιτροπή δεν γνωρίζει πόσοι αγοραστές έχουν ανοίξει λογαριασμούς για πληρωμές προμήθειας φυσικού αερίου μέσω της \en Gazprombank. \el Στο μεταξύ, ανώτερος αξιωματούχος της Ευρωπαϊκής Ένωσης δήλωσε πως και μόνο το άνοιγμα τραπεζικού λογαριασμού σε ρούβλια στην \en Gazprombank \el ενδέχεται να αποτελεί παραβίαση των κυρώσεων που έχει επιβάλει η ΕΕ σε βάρος της Ρωσίας, όμως η ΕΕ δεν έχει ένδειξη πως κάποια εταιρεία φυσικού αερίου της ΕΕ έχει κάνει κάτι τέτοιο. Ο αξιωματούχος δήλωσε πως «εκ πρώτης όψεως» το άνοιγμα τραπεζικών λογαριασμών σε ρούβλια από εισαγωγείς φυσικού αερίου φαίνεται ότι παραβιάζει τις κυρώσεις. Ο αξιωματούχος πρόσθεσε πως η Ευρωπαϊκή Επιτροπή δεν έχει κάποια επίσημη ένδειξη ότι εταιρείες της ΕΕ έχουν δημιουργήσει στην\en Gazprombank \el λογαριασμούς σε ρούβλια για την πληρωμή του φυσικού αερίου. Επίσης διευκρίνισε πως η Πολωνία και η Βουλγαρία χρησιμοποίησαν τις υφιστάμενες μεθόδους πληρωμής για το ρωσικό αέριο, πριν η Μόσχα αναστείλει χθες, Τετάρτη, τις προμήθειες των χωρών αυτών με αέριο, και πως δεν χρησιμοποίησαν τον μηχανισμό που προτείνει η Μόσχα για να πληρώσουν σε ρούβλια. «Σύμφωνα με τις πληροφορίες μας, αμφότερες οι χώρες επέμειναν στην αρχική μορφή πληρωμής», δήλωσε ο αξιωματούχος σε δημοσιογράφους. Ωστόσο δύο πηγές είπαν σήμερα στο Ρόιτερς ότι λίγες ευρωπαϊκές εταιρείες έχουν αρχίσει να πληρώνουν σε ρούβλια τη Ρωσία για το φυσικό αέριο, αν και μεγάλοι πελάτες της δεν το έχουν κάνει ακόμη. «Μερικές εμπορικές εταιρείες, ίσως περισσότερες από πέντε, έχουν αρχίσει τις πληρωμές», είπε μία πηγή, ζητώντας να μην κατονομαστεί, επειδή δεν είχε εξουσιοδοτηθεί να μιλήσει στα μέσα ενημέρωσης. \\ 
     \hline 
     \hline 
     \multirow{5}{*}[-3.5em]{\rotatebox[origin=c]{90}{\textsc{Title}}} & Gold &  \selectlanguage{greek} Φυσικό αέριο: Όλες οι χώρες της ΕΕ συμφωνούν ότι δεν θα πληρώσουν τη Ρωσία σε ρούβλια\\ 
     & BART-random &\selectlanguage{greek} Ε.Ε.: «Δεν θα πληρώσουν» οι χώρες της ΕΕ για το φυσικό αέριο σε ρούβλια\\ 
     & mBART25 &\selectlanguage{greek} ΕΕ: Οι χώρες δεν πληρώνουν σε ρούβλια τη Ρωσία για το φυσικό αέριο\\
     & mBART50 &\selectlanguage{greek} ΕΕ: Οι χώρες δεν πληρώνουν σε ρούβλια τη Ρωσία για το φυσικό αέριο\\
     & GreekBART &\selectlanguage{greek} ΕΕ: Δεν θα πληρώσουμε τη Ρωσία σε ρούβλια για το φυσικό αέριο\\ 
     \hline 
     \end{tabular} 
     \caption{Example 2-GreekSUM Title} 
     \label{appendix:table2t} 
     \end{table}

\clearpage
 
  \begin{table}[h!] 
     \centering 
     \begin{tabular}{|cr|p{5.5in}|}
     \hline 
     & Document & \el Στις ημέρες του Πάσχα έχει προσαρμοστεί το πρόγραμμα λειτουργίας λεωφορείων, τρόλεϊ, ηλεκτρικού και μετρό. Ειδικότερα, τα λεωφορεία και τα τρόλεϊ σήμερα, Μεγάλη Παρασκευή, θα κινούνται με πρόγραμμα Σαββάτου. Οι συρμοί στο μετρό θα διέρχονται από τους σταθμούς ανά 7 λεπτά από τις 09.00 έως τις 17.00 και ανά 10 λεπτά τις υπόλοιπες ώρες. Υπενθυμίζεται πως δεν θα ισχύσει η δίωρη παράταση λειτουργίας που εφαρμόζεται τις Παρασκευές. Στον ηλεκτρικό οι συρμοί θα διέρχονται από τους σταθμούς ανά 10,5 λεπτά. Τα λεωφορεία και τα τρόλεϊ θα κινηθούν με πρόγραμμα Κυριακής, ενώ θα αποσυρθούν νωρίτερα, ώστε να βρίσκονται στα αμαξοστάσια στις 23.00. Τα λεωφορεία θα κινηθούν με πρόγραμμα Κυριακής και τα τρόλεϊ με ειδικό πρόγραμμα Κυριακής. Τόσο στα δρομολόγια των λεωφορείων όσο και σ' αυτά των τρόλεϊ θα εφαρμοστεί ειδικό πρόγραμμα Σαββάτου. Ακινητοποιημένοι θα μείνουν την Τετάρτη 1η Μαΐου οι συρμοί του ηλεκτρικού (πρώην ΗΣΑΠ), τα λεωφορεία, τα τρόλεϊ, αλλά και ο σιδηρόδρομος, λόγω 24ωρης απεργίας των εργαζομένων, που θα συμμετάσχουν στις απεργιακές συγκεντρώσεις για την Πρωτομαγιά. Όπως αναφέρουν σε ανακοίνωσή τους οι εργαζόμενοι στον πρώην ΗΣΑΠ, «είναι μέρα αγώνα, τιμής και μνήμης. Θυμόμαστε και τιμάμε τους πρωτοπόρους αγωνιστές και τα θύματα των εργατικών αγώνων για βελτίωση των συνθηκών δουλειάς για αξιοπρεπείς αμοιβές και την κατοχύρωση των δικαιωμάτων μας. Ανασυγκροτούμαστε, θέτουμε τους στόχους μας και προχωράμε σε νέους αγώνες. Διεκδικούμε και παλεύουμε για την αναπλήρωση απωλειών από τις μνημονιακές πολιτικές λιτότητας, για πραγματικές αυξήσεις στους μισθούς και στις κοινωνικές παροχές». Και προσθέτουν «υπερασπιζόμαστε τον δημόσιο χαρακτήρα των συγκοινωνιών. Διεκδικούμε την υπογραφή νέας Συλλογικής Σύμβασης Εργασίας. Αγωνιζόμαστε για ασφαλείς, φθηνές συγκοινωνίες. Με αγώνες κατακτάμε τα δικαιώματά μας». \\ 
     \hline 
     \hline 
     \multirow{5}{*}[-3.5em]{\rotatebox[origin=c]{90}{\textsc{Title}}} & Gold &  \selectlanguage{greek}Πάσχα 2019: Πώς θα κινηθούν λεωφορεία, τρόλεϊ, ηλεκτρικός και μετρό\\ 
     & BART-random &\selectlanguage{greek}Μέσα Μαζικής Μεταφοράς: Πώς θα κινηθούν σήμερα τα Μέσα Μεταφοράς \\ 
     & mBART25 &\selectlanguage{greek}Μέσα Πάσχα: Πώς θα κινηθούν σήμερα λεωφορεία, τρόλεϊ, ηλεκτρικό και μετρό \\
     & mBART50 &\selectlanguage{greek}Μέσα Πάσχα: Πώς θα κινηθούν σήμερα λεωφορεία, τρόλεϊ, ηλεκτρικό και μετρό \\
     & GreekBART &\selectlanguage{greek}Πάσχα: Πώς θα κινηθούν σήμερα λεωφορεία, τρόλεϊ, ηλεκτρικού και μετρό\\ 
     \hline 
     \end{tabular} 
     \caption{Example 3-GreekSUM Title} 
     \label{appendix:table3t} 
     \end{table}

\clearpage

    \begin{table}[h!] 
     \centering 
     \begin{tabular}{|cr|p{5.5in}|}
     \hline 
     & Document & \el Συνάντηση με οικονομικούς παράγοντες από το Σίτι του Λονδίνου έχει αυτή την ώρα ο Αλέξης Τσίπρας στο κέντρο της βρετανικής πρωτεύουσας. Τον Έλληνα πρωθυπουργό υποδέχθηκε ο αντιπρόεδρος της Επιτροπής Πολιτικής του Σίτι, Τομ Σλέι \en (Tom Sleigh). \el Επισημαίνεται ότι η Επιτροπή υπέχει θέση Διοίκησης του Σίτι του Λονδίνου. Από την αίθουσα της «Παλιάς Βιβλιοθήκης», ο πρωθυπουργός θα απευθυνθεί σε έναν κύκλο περισσότερων από εκατό σημαίνοντων στελεχών της επενδυτικής/χρηματοπιστωτικής κοινότητας του Σίτι και, σύμφωνα με πληροφορίες, στη συνέχεια θα ακολουθήσει συνάντηση σε πιο στενό κύκλο συμμετεχόντων. Στον απόηχο της απόφασης του \en Eurogroup \el για την ελάφρυνση του χρέους, οι επαφές του Αλέξη Τσίπρα με σημαντικούς εκπροσώπους της επενδυτικής/χρηματοπιστωτικής της κοινότητας του οικονομικού κέντρου της Ευρώπης, σηματοδοτούν ένα ευκρινές διεθνές μήνυμα για τις προοπτικές της ελληνικής οικονομίας και της «επόμενης μέρας», στην περίοδο μετά την ολοκλήρωση των μνημονίων. Όπως ανέφερε κυβερνητικός αξιωματούχος, οι σημερινές συναντήσεις είναι ένας σημαντικός σταθμός σε μια «αλυσίδα» επαφών και συνομιλιών που θα συνεχιστούν στο αμέσως επόμενο διάστημα των καλοκαιρινών μηνών και το φθινόπωρο. Ενδεικτική της ευνοϊκής συγκυρίας για την ελληνική οικονομία και το στοίχημα της ανάκαμψης, η χθεσινοβραδινή αναβάθμιση, από τον αμερικανικό οίκο αξιολόγησης \en Standard \& Poor's \el της μακροπρόθεσμης πιστοληπτικής ικανότητας της χώρας σε \en B+, \el χαιρετίζοντας την απόφαση του \en Eurogroup \el. Στις 18:00 το απόγευμα ώρα Ελλάδας, ο πρωθυπουργός θα περάσει το κατώφλι της \en Downing Street 10 \el προκειμένου να συναντηθεί με την πρωθυπουργό της Βρετανίας, Τερέζα Μέι. Στη συνέχεια θα έχει συνάντηση με τον αρχηγό του Εργατικού Κόμματος, Τζέρεμι Κόρμπιν. \\ 
     \hline 
     \hline 
     \multirow{5}{*}[-3.5em]{\rotatebox[origin=c]{90}{\textsc{Title}}} & Gold &  \selectlanguage{greek}Συνάντηση με οικονομικούς παράγοντες από το Σίτι του Λονδίνου έχει ο Αλέξης Τσίπρας \\ 
     & BART-random &\selectlanguage{greek}Μήνυμα Τσίπρα στο Λονδίνο για το χρέος \\ 
     & mBART25 &\selectlanguage{greek}Συνάντηση Τσίπρα με οικονομικούς παράγοντες στο Σίτι \\
     & mBART50 &\selectlanguage{greek}Συνάντηση Τσίπρα με οικονομικούς παράγοντες στο Σίτι \\
     & GreekBART &\selectlanguage{greek}Βλέμματα στο Λονδίνο για την ελληνική οικονομία\\ 
     \hline 
     \end{tabular} 
     \caption{Example 4-GreekSUM Title} 
     \label{appendix:table4t} 
     \end{table}

\clearpage

     \begin{table}[h!] 
     \centering 
     \begin{tabular}{|cr|p{5.5in}|}
     \hline 
     & Document & \el Επιβατικό τρένο εκτροχιάστηκε σήμερα περίπου 20 χλμ. βόρεια της Ραμπάτ, με αποτέλεσμα να σκοτωθούν έξι άνθρωποι και άλλοι 86 να τραυματιστούν, σύμφωνα με επίσημο απολογισμό που ανακοινώθηκε στον τόπο του δυστυχήματος. «Ο εκτροχιασμός προκάλεσε έξι θανάτους, σύμφωνα με τον τρέχοντα απολογισμό, και 86 τραυματίες σε σοβαρή κατάσταση», δήλωσε ο Μοχάμεντ Ραμπί Ραχίλ, γενικός διευθυντής της εταιρίας σιδηροδρόμων \en ONCF, \el ο οποίος μετέβη επί τόπου. «Ξεκίνησε έρευνα για τον προσδιορισμό των αιτιών του δυστυχήματος», πρόσθεσε, σε βίντεο που αναρτήθηκε στα μέσα κοινωνικής δικτύωσης. Θεαματικές εικόνες του δυστυχήματος, που σημειώθηκε γύρω στις 13:00 ώρα Ελλάδας, περίπου 20 χλμ. βόρεια της πρωτεύουσας Ραμπάτ, στο ύψος της κοινότητας Σιντί Μπουκναντέλ, κάνουν τον γύρο των μέσων κοινωνικής δικτύωσης, που είναι πολύ επικριτικά εναντίον της \en ONCF. \el Οι εικόνες δείχνουν πολλά βαγόνια εκτροχιασμένα κοντά σε μια γέφυρα στους αγρούς, ενώ η μηχανή είναι πλήρως κατεστραμμένη. Ο οδηγός της αμαξοστοιχίας είναι νεκρός, σύμφωνα με πολλά τοπικά ΜΜΕ. Ο βασιλιάς αποφάσισε να αναλάβει τα έξοδα της κηδείας των θυμάτων και οι τραυματίες θα διακομιστούν στο στρατιωτικό νοσοκομείο της Ραμπάτ με βασιλικές οδηγίες, αναφέρεται σε ανακοίνωση του γραφείου του βασιλιά. \\ 
     \hline 
     \hline 
     \multirow{5}{*}[-3.5em]{\rotatebox[origin=c]{90}{\textsc{Title}}} & Gold &  \selectlanguage{greek}Εκτροχιασμός τρένου στο Μαρόκο: Στους 6 οι νεκροί - 86 τραυματίες \\ 
     & BART-random &\selectlanguage{greek}Ραμπάτ: 20 νεκροί από εκτροχιασμό τρένου\\ 
     & mBART25 &\selectlanguage{greek}ΗΠΑ: Επιβατικό τρένο εκτροχιάστηκε - Έξι νεκροί και 86 τραυματίες \\
     & mBART50 &\selectlanguage{greek}ΗΠΑ: Επιβατικό τρένο εκτροχιάστηκε - Έξι νεκροί και 86 τραυματίες\\
     & GreekBART &\selectlanguage{greek}Εκτροχιασμός τρένου στη Ραμπάτ: Έξι νεκροί και 86 τραυματίες \\ 
     \hline 
     \end{tabular} 
     \caption{Example 5-GreekSUM Title} 
     \label{appendix:table5t} 
     \end{table}

\end{document}